\documentclass[10pt,twocolumn,letterpaper]{article}

\usepackage{graphicx}
\usepackage{amsmath}
\usepackage{amssymb}
\usepackage{booktabs}

\makeatletter
\renewcommand{\maketag@@@}[1]{\hbox{\m@th\normalsize\normalfont#1}}%
\makeatother
\usepackage{multirow}
\usepackage{array}
\usepackage[table]{xcolor}
\usepackage{tablefootnote}
\usepackage{multicol}

\usepackage[pagebackref,breaklinks,colorlinks]{hyperref}

\usepackage[capitalize]{cleveref}
\crefname{section}{Sec.}{Secs.}
\Crefname{section}{Section}{Sections}
\Crefname{table}{Table}{Tables}
\crefname{table}{Tab.}{Tabs.}

\usepackage{hyperref}
\hypersetup{colorlinks,allcolors=red}

\def\cvprPaperID{295} 
\def\confName{CVPR}
\def\confYear{2023}

\begin{document}

\title{Self-distillation with Online Diffusion on Batch Manifolds Improves Deep Metric Learning}

\author{Zelong Zeng\\
The University of Tokyo\\
\& National Institute of Informatics\\
{\tt\small zzlbz@nii.ac.jp}
\and
Fan Yang\\
National Institute of Informatics\\
{\tt\small yang@nii.ac.jp}
\and
Hong Liu\\
National Institute of Informatics\\
{\tt\small hliu@nii.ac.jp}
\and
Shin'ichi Satoh\\
National Institute of Informatics\\
\& The University of Tokyo\\
{\tt\small satoh@nii.ac.jp}
}

\maketitle

\begin{abstract}
Recent deep metric learning (DML) methods typically leverage solely class labels to keep positive samples far away from negative ones.
However, this type of method normally ignores the crucial knowledge hidden in the data (e.g., intra-class information variation), which is harmful to the generalization of the trained model. 
To alleviate this problem, in this paper we propose Online Batch Diffusion-based Self-Distillation (OBD-SD) for DML. 
Specifically, we first propose a simple but effective Progressive Self-Distillation (PSD), which distills the knowledge progressively from the model itself during training. 
The soft distance targets achieved by PSD can present richer relational information among samples, which is beneficial for the diversity of embedding representations.
Then, we extend PSD with an Online Batch Diffusion Process (OBDP), which is to capture the local geometric structure of manifolds in each batch, so that it can reveal the intrinsic relationships among samples in the batch and produce better soft distance targets.
Note that our OBDP is able to restore the insufficient manifold relationships obtained by the original PSD and achieve significant performance improvement. 
Our OBD-SD is a flexible framework that can be integrated into state-of-the-art (SOTA) DML methods.  
Extensive experiments on various benchmarks, namely CUB200, CARS196, and Stanford Online Products, demonstrate that our OBD-SD consistently improves the performance of the existing DML methods on multiple datasets with negligible additional training time, achieving very competitive results. Code: \url{https://github.com/ZelongZeng/OBD-SD_Pytorch}
\end{abstract}

\section{Introduction}
\label{sec:intro}
Deep Metric Learning (DML) aims to learn an embedding space that uses predefined distance metrics (\textit{e.g.}, Euclidean distance) to reasonably measure the similarity between training samples and then transfer to unseen test data. 
The generalization capability of embedding space is very essential for many downstream applications of DML, such as image retrieval~\cite{gordo2016deep, gordo2017end}, face recognition~\cite{schroff2015facenet, deng2019arcface}, person re-identification~\cite{wang2015zero, hermans2017defense} and representation learning~\cite{wang2020cross}. 

Most existing DML methods utilize a pre-defined distance metric, depending on the class label, to increase the inter-class distance in the embedding space while decreasing the intra-class distance \cite{suarez2021tutorial}. 
Typically, these methods promote intra-class samples to form a compact cluster with a large margin from other clusters, thus learning a strongly discriminative representation for the seen class, but ignoring the intrinsic relationships of the samples, such as intra-class variation. 
In particular, since the relationships of samples in the embedding space are derived solely from the distance metric based on the class labels, the diversity of the relationships of samples in the embedding space is suppressed.
However, ignoring the intrinsic relationships of the samples actually harms the generalization of learned embedding space since they are useful to identify unseen classes~\cite{sanakoyeu2019divide,milbich2020diva,xuan2020improved,zheng2021deep,zheng2021deepcompos}.
 


To enhance this diversity, we propose a simple yet effective method called \textit{Online Batch Diffusion-based Self-distillation} (OBD-SD), which is composed of two complementary components. 
Specifically, inspired by recent typical knowledge distillation techniques~\cite{furlanello2018born, yun2020regularizing, kim2021self,yuan2020revisiting}, we first propose Progressive Self-distillation (PSD) that can be integrated into many typical DML methods \cite{wu2017sampling,wang2019multi,roth2020revisiting}. 
For PSD, we define the student model trained in the previous epoch as the teacher model, which means that the student model serves as its own teacher model. 
This procedure does not require any modifications to the network architecture and incurs no further computational cost. 
In other words, in a dynamic manner, PSD distills the distance metric from the model at the previous epoch and then adaptively assigns a soft distance target for each training pair of data. 
We theoretically and experimentally analyze that the soft distance targets learned from PSD have more information about how the training samples are related to each other (rather than only considering the class label information). This helps to learn more diverse embedding representations and solves the problem described above.


It's worth noting that a better distance metric should consider the local geometric structure of manifolds, rather than just using the rigid distance metrics, which is disregarded by our proposed PSD. 
Therefore, inspired by existing diffusion process methods~\cite{zhou2003ranking, bai2017regularized, yang2019efficient}, we use the diffusion process to refine the distance target provided by the teacher model of PSD.
We further proposed an Online Batch Diffusion Process (OBDP) technique, which aims to capture the local geometry structure of manifolds in a mini-batch. 
Our OBDP is able to learn the high-order information from batch manifolds, which presents the intrinsic relationships between samples and makes the soft distance targets more accurate. 
Note that, our OBDP makes use of a mini-batch to construct the manifold structure, which belongs to the online learning scheme, thereby significantly enhancing the computational efficiency.
Our experiments and observations show that, compared with the existing methods doing diffusion on the global manifolds, this online learning scheme can achieve competitive or even better results. 

We reformulate our OBD-SD model combing PSD and OBDP.
The OBD-SD is a plug-and-play technique that can be integrated into most conventional DML methods with negligible extra computation cost.  
Our theoretical studies (as presented in Section~\ref{sec:theoretical}) reveal that our OBD-SD can take into account the high-order relationship of samples, especially for hard samples.
That is, our OBD-SD can learn a more accurate distance metric, which benefits the overall performance.



Finally, we conduct extensive experiments on three widely-used DML benchmarks, \emph{i.e,} CUB200-2011~\cite{wah2011caltech}, CARS196~\cite{krause20133d} and Stanford Online Products~\cite{oh2016deep}, on which the results validate the merits of the proposed OBD-SD over the state-of-the-art (SOTA) DML methods. 
Moreover, experiments in Section~\ref{subsection:space_metrics} indicate that our OBD-SD indeed makes the learned embedding features achieve higher diversity and mitigate over-clustering~\cite{roth2021simultaneous}, both encouraging better generalization to unseen test classes.

\section{Related Works}
\label{sec:related}

\subsection{Deep Metric Learning}

Metric learning aims to learn a good metric space, in which we can measure the similarity between samples by using distance metrics. 
Recent works leverage deep learning techniques to facilitate a better distance metric, which is called deep metric learning (DML).
The pioneering works consider the relationship between a tuple of samples, such as pairs~\cite{hadsell2006dimensionality}, triplets~\cite{wang2014learning,schroff2015facenet,wu2017sampling}, and other variants (like N-pairs~\cite{sohn2016improved} and lifted structured loss~\cite{oh2016deep}). 
Recently, AP-based losses~\cite{brown2020smooth,rolinek2020optimizing} and proxy-based methods~\cite{movshovitz2017no,qian2019softtriple,kim2020proxy} have demonstrated the capacity to learn distance-preserving embedding spaces. 
In addition, many different sampling schemes have been introduced to reduce the computational complexity caused by a large number of training pairs ~\cite{schroff2015facenet,wu2017sampling,ge2018deep,roth2019mic,roth2020pads}. 

Recently, a promising direction is to explore how to learn a discriminative embedding space with good generalization to unseen classes. 
Most existing methods of this type consider learning an ensemble of embeddings by manually designing different learning objectives or constraints and demonstrate that the diversity of embedding representations is crucial to the generalization of embedding space~\cite{sanakoyeu2019divide,roth2019mic,milbich2020diva,xuan2020improved,zheng2021deep,zheng2021deepcompos}. However, they usually face the problem of less scalability and high computational complexity.
In contrast, our OBD-SD is efficient and straightforward, showing good scalability and generalization capabilities.

\subsection{Knowledge Distillation}

Early knowledge distillation (KD) attempts to compress models by transferring the ``dark knowledge'' from the larger teacher model to a compact student model~\cite{hinton2015distilling,zagoruyko2016paying,tung2019similarity}. 
Many recent works find that the student model can outperform the teacher model when the student is configured with the same capacity as its teacher~\cite{furlanello2018born,yuan2020revisiting}. Based on this observation, they introduce a novel self-distillation (Self-KD) scheme. 
In addition, some other works, such as PS-KD~\cite{kim2021self}, BYOT~\cite{zhang2019your}, and CS-KD~\cite{yun2020regularizing}, present a more effective and efficient simultaneous Self-KD scheme.

More recently, researchers begin to apply the Self-KD for improving the generalization of DML, taking RCL~\cite{kim2021embedding}, BAR~\cite{qin2021born}, and S2SD~\cite{roth2021simultaneous} as representative works. 
BAR applies the conventional strategy, \textit{i.e.}, using a pre-trained model as the teacher. And S2SD extends DML with knowledge distillation from multiple auxiliary heads with high-dimension, all auxiliary heads share the same backbone with the base networks so they can be trained simultaneously.
It's worth noting that our work is somewhat similar to S2SD. We elaborate on some major differences between S2SD and ours as follows:
First, S2SD targets the task of ``feature compression'', \textit{i.e.}, using high-dimensional embedding space with better generalization capacity to assist the training of low-dimensional embedding space. 
In contrast, Our OBD-SD proposes that the same dimensional embedding (as opposed to the high-dimensional embedding of S2SD) space itself can provide helpful information to improve performance, while S2SD does not. 
Second, our OBD-SD proposes Online Batch Diffusion Process (OBDP) for refining soft distance targets from the teacher. We also apply our OBDP to the existing Self-KD-based methods, and the experiments in Section~\ref{subsec:comparision_self-kd} show that our OBDP can significantly improve the performance of the compared Self-KD-based DML methods. 

\subsection{Diffusion Process}

The diffusion process is commonly used as a re-ranking method in retrieval tasks~\cite{zhou2003ranking,donoser2013diffusion,bai2017regularized,iscen2017efficient,yang2019efficient}. 
Due to the diversity of data, there is usually a distribution difference between training data and unknown test data \cite{shalev2014understanding}. 
The difference often results in manifolds in the embedding space that are not conducive to measuring the relationships between test samples using the rigid distance metric~\cite{yang2019efficient}. 
To handle this problem, diffusion process can capture the local geometry structure of the data manifolds based on a neighborhood graph, which can make distance metric more accurately~\cite{bai2017regularized}.
Such a neighborhood graph consists of various nodes and edges, where each node represents a feature from the database and the edge presents the connection between each node and its  neighborhood with corresponding weights proportional to the pairwise affinities between nodes. 
Additionally, diffusion process is often applied in label propagation for deep semi-supervised learning~\cite{zhou2003learning,iscen2019label,huang2020combining}. 
Different from these diffusion process methods proposed as the post-processing method, our OBD-SD uses diffusion process as an auxiliary module that can refine the soft distance targets well during the training.
Moreover, our OBD-SD uses only the mini-batch data for each diffusion process, which is more efficient than the conventional methods of using global data.

\begin{figure*}[t]
\centering
\includegraphics[width=0.8\linewidth]{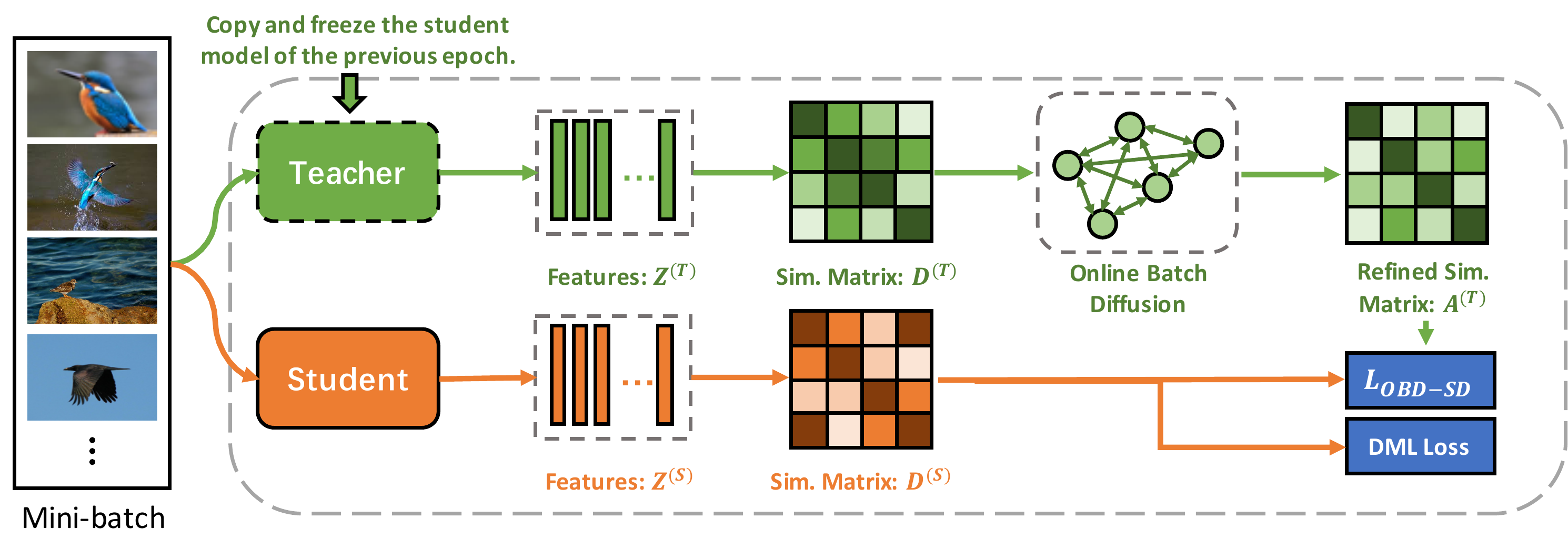}
\caption{A schematic diagram of OBD-SD. The student trained in the previous epoch is the teacher in the current epoch. 
During the training, for each mini-batch, we use the teacher and the student models to produce the similarity matrices $D^{(T)}$ and $D^{(S)}$, respectively. Then, use online batch diffusion to refine $D^{(T)}$ to produce the refined similarity matrix $A^{(T)}$. Finally, compute DML losses while applying OBD-SD loss function on the matrices $A^{(T)}$ and $D^{(S)}$.}
\label{fig:framework}
\end{figure*}

\section{Proposed Method}
\label{sec:approach}


\subsection{Problem Formulation}

For metric learning, we denote $\mathcal{X} = \left \{ \boldsymbol{x}_{1},\boldsymbol{x}_{2},...,\boldsymbol{x}_{n} \right \}$ as the training data, and $y_i$ is the corresponding label of $\boldsymbol{x}_i$. 
Our goal is to learn an embedding function $f:\mathcal{X} \mapsto \Phi \subset \mathbb{R}^d$ that maps each input $\boldsymbol{x}\in\mathcal{X}$ to the $d$-dimensional embedding space $\Phi$ that allows measuring the similarity between inputs. 
To this end, Let $f$ be formulated as a differentiable deep network $f(\cdot,\theta)$ with a trainable parameter $\theta$. The output $\boldsymbol{v}_{i} = f(\boldsymbol{x}_{i},\theta)$ is typically normalized to $\boldsymbol{z}_{i} = \frac{\boldsymbol{v}_{i}}{\left \| \boldsymbol{v}_{i} \right \|}$. 
The distance (or similarity) between two inputs $\boldsymbol{x}_{i}, \boldsymbol{x}_{j}$ in the learned metric space is defined as  $D_{i,j}:=d(\boldsymbol{z}_{i},\boldsymbol{z}_{j})$, 
where $d(\cdot,\cdot)$ is the pre-defined distance (or similarity) function. 
In this paper, we use cosine similarity as the default, \textit{i.e.}, 
$D_{i,j}={\boldsymbol{z}_{i}}\cdot\boldsymbol{z}_{j}$.

\subsection{Progressive Self-distillation}

In this paper, we aim to solve the problem of ignoring the intrinsic relationships, so that a feasible way is to assign a more precise distance target for each pair. 
However, we usually only have the class labels of the data with the lack of characteristic annotations (e.g., attribute labels), which makes current DML difficult to get actual distance targets only by the annotation of the samples. 
Inspired by previous works~\cite{furlanello2018born, qin2021born}, we exploit knowledge distillation strategy to obtain soft distance targets from a teacher model. 

To achieve this goal, the first problem we need to solve is how to obtain the teacher model. 
Previous works usually use a pre-trained model~\cite{qin2021born, kim2021embedding} or add auxiliary architecture ~\cite{roth2021simultaneous} as the teacher model.
However, these methods are inflexible and require a longer training time. 

An interesting observation~\cite{yuan2020revisiting} is that a relatively poorly-trained teacher model can still help improve the student model. 
Inspired by this, we proposed a Progressive Self-distillation (PSD), which dynamically exploits the model trained in the previous epoch as the teacher model. 
In special, when finishing the training of the model in epoch $(t-1)$, we copy and freeze this model as the teacher model used in the next epoch, which is defined as $f^{(T)}$. 
When we continue to train the student model $f^{(S)}$ in the epoch $(t)$, we use the $f^{(T)}$ to provide soft distance targets for knowledge distillation. 

Then, the second problem is how to design the knowledge distillation loss function. 
Similar to previous works~\cite{qin2021born, roth2021simultaneous}, PSD matches sample relationships within a mini-batch $\mathcal{B}$ in base space and target space, which can ensure the generality of the KD functions while fully utilizing the information of samples in the batch. 
Specifically, we calculate the similarity matrix $D\in\mathbb{R}^{|\mathcal{B}|\times|\mathcal{B}|}$, where $D_{i,j}$ indicates the cosine similarity between $i$-th and $j$-th sample in the mini-batch $\mathcal{B}$. 
Then, we use KL divergence as the basic of our PSD loss, which is shown as:
\begin{equation}
    L_{\textup{PSD}}(\mathcal{B}) = \frac{1}{\left | \mathcal{B} \right|}
    \displaystyle\sum_{i=1}^{|\mathcal{B}|}
    \displaystyle\sum_{j=1}^{|\mathcal{B}|}\sigma(D_{i,:}^{(T)}/\tau)_j\textup{ln}\frac{\sigma(D_{i,:}^{(T)}/\tau)_j}{\sigma(D_{i,:}^{(S)}/\tau)_j},
    \label{eq:psd}
\end{equation}
where $\sigma()$ is the softmax operation, $\tau$ is a temperature parameter, $D_{i,:}$ indicates the $i$-th row of the matrix $D$, and $D^{(T)}$ and $D^{(S)}$ are the similarity matrix of $f^{(T)}$ and $f^{(S)}$, respectively. 

However, the teacher model cannot learn sufficient knowledge from the training data in early epochs of training but becomes increasingly reliable as the epoch grows.
Considering this, we use a dynamic weight updating scheme~\cite{kim2021self} to change the weight of PSD loss online. 
The dynamic weight formulation is $\frac{t}{T}\lambda$, where $\lambda$ is the final weight for distillation, $t$ is the current epoch, and $T$ is the total epoch of training. 
We find that this dynamic updating scheme can reach a better performance than static weight. 
Finally, we extend the typical DML with our PSD, whose final objective function is written as:
\begin{equation}
    \mathcal{L} = L_{\textrm{DML}} + \tau^2\frac{t}{T}\lambda L_{\textrm{PSD}},
\label{eq:dml_psd}
\end{equation}
where $L_{\textrm{DML}}$ is an arbitrary metric learning loss function and $\tau^2$ is a balanced parameter often applied in KD loss~\cite{hinton2015distilling}.

\subsection{Online Batch Diffusion Process}
\label{sec:OBDP}

However, there is another key problem in the PSD scheme, \emph{i.e.,} the soft distance targets from the teacher model cannot fully reflect the intrinsic relationships among samples, because of the limitation of the rigid distance metrics~\cite{bai2017regularized,yang2019efficient}.
To solve this problem, we proposed Online Batch Diffusion Process (OBDP) for refining the soft distance targets from the teacher model. OBDP can help the embedding space learn more about intrinsic relationships among samples and improve performance.

\subsubsection{Preliminaries of Diffusion Process}
\label{subsec:preliminaries}
We first revisit the standard diffusion process in image retrieval (mainly following the steps from~\cite{zhou2003ranking, yang2019efficient}). 
Given a data set $\mathcal{U}=\{ u_1,\dots, u_q, u_{q+1},\dots,u_{n}\}$, the first $q$ samples are the queries and the rest are the gallery. 
We define $\mathbf{f}^0=[\mathbf{f}_1^0,\dots,\mathbf{f}_n^0]^T$, in which $\mathbf{f}_i^0=1$ if $u_i$ is a query and $\mathbf{f}_i^0=0$ otherwise. The typical diffusion process has three steps, which can be described as follows:
\begin{itemize} 
    \item [1.] Use a pre-trained model to extract the features $\mathcal{Z}=\{\boldsymbol{z}_{1},\dots,\boldsymbol{z}_n\}$ from dataset $\mathcal{U}$ and calculate the affinity matrix $W\in\mathbb{R}^{n\times n}$, where each component $W_{ij} = d(\boldsymbol{z}_{i},\boldsymbol{z}_{j})$. Note that $W_{ii} = 0$ because there are no loops in the graph.
    \item [2.] Symmetrically normalize $W$ by $S=V^{-1/2}WV^{-1/2}$ $\in\mathbb{R}^{n\times n}$, in which $V\in\mathbb{R}^{n\times n}$ is the diagonal matrix with $V_{ii}=\sum_{j=1}^{n}W_{ij}$. We refer to $S$ as transition matrix.
    \item [3.] Use random walk to iterate $\mathbf{f}^{t+1}=\omega S \mathbf{f}^{t}+(1-\omega)\mathbf{f}^0$ until convergence, where $\omega\in(0,1)$ is the transition probability of random walk.
\end{itemize}

It can be proved that the above iteration converges to a closed-form solution: 
\begin{equation}
    \mathbf{f}^{\infty}=(1-\omega)(I-\omega S)^{-1}\mathbf{f}^{0}\in\mathbb{R}^{n}. 
\label{eq:diffusion}
\end{equation}
where $\mathbf{f}^{\infty}$ can be used as ranking scores for re-ranking. 

\subsubsection{Online Diffusion on Batch Manifolds}
The standard diffusion process is commonly used as a re-ranking method. 
Though the Eq~\ref{eq:diffusion} requires $O(N^3)$ time complexity ($N$ is the number of samples), we only need to compute it once offline. 
However, if we want to use diffusion process during training, we need to update and maintain the Eq~\ref{eq:diffusion} online, which is impractical since the size of the training set is large. 

To solve this problem, we proposed Online Batch Diffusion Process (OBDP), in which we only use the samples in the mini-batch to construct the neighborhood graph for diffusion process each time. 
This means that we only diffuse information on batch manifolds. 
Specifically, for each mini-batch $\mathcal{B}$, we first use a teacher model to extract the features $\mathcal{Z}^{(T)}\!\in\mathbb{R}^{|\mathcal{B}|\times d}$ and then calculate the similarity matrix $D\in\mathbb{R}^{|\mathcal{B}|\times |\mathcal{B}|}$. 
Note that since our aim is to refine relationships between samples rather than simply calculate ranking scores, we use $D$ as the initial state for diffusion in OBDP, which is different from the standard diffusion process.
Then we calculate the affinity matrix $W\in\mathbb{R}^{|\mathcal{B}|\times|\mathcal{B}|}$ and transition matrix $S\in\mathbb{R}^{|\mathcal{B}|\times|\mathcal{B}|}$ based on the extracted $\mathcal{Z}^{(T)}$. 
Finally, we use random walk to refine the initial states $D$:
\begin{equation}
    A=(1-\omega)(I-\omega S)^{-1}D\in\mathbb{R}^{|\mathcal{B}|\times |\mathcal{B}|},\ \omega \in (0,1) 
\label{eq:OBDP_2}
\end{equation}
where $A$ is the refined similarity matrix of the mini-batch $\mathcal{B}$. Since $|\mathcal{B}|$ is usually much smaller than the size of training set $N$, OBDP can be efficiently computed online. Moreover, if $|\mathcal{B}|$ is fixed, Eq~\ref{eq:OBDP_2} only requires $O(N)$ for each epoch, which is more efficient than the standard diffusion process.

\subsection{Online Batch Diffusion-based Self-distillation}

Now, we can use OBDP to refine the soft distance targets from PSD. Specifically, for each mini-batch $\mathcal{B}$, we use the teacher model $f^{(T)}$ to extract the features $\mathcal{Z}^{(T)}$ and apply OBDP to calculate the refined similarity matrix $A^{(T)}$:
\begin{equation}
    A^{(T)}=(1-\omega)(I-\omega S^{(T)})^{-1}D^{(T)},
\label{eq:OBD-SD_1}
\end{equation}
where the similarity matrix $D^{(T)}$ and transition matrix $S^{(T)}$ are both calculated by using $\mathcal{Z}^{(T)}$. Then, we replace $D^{(T)}$ in Eq~\ref{eq:psd} with $A^{(T)}$. Thus our OBD-SD loss is:
\begin{equation}
    L_{\textup{OBD-SD}}(\mathcal{B})\!=\!\frac{1}{\left | \mathcal{B} \right|}
    \displaystyle\sum_{i=1}^{|\mathcal{B}|}
    \displaystyle\sum_{j=1}^{|\mathcal{B}|}\sigma(A_{i,:}^{(T)}\!/\tau)_j\textup{ln}\frac{\sigma(A_{i,:}^{(T)}\!/\tau)_j}{\sigma(D_{i,:}^{(S)}\!/\tau)_j},
\label{eq:obdsd}
\end{equation}
Overall, the final objective function can be described as:
\begin{equation}
    \mathcal{L} = L_{\textrm{DML}} + \tau^2\frac{t}{T}\lambda L_{\textrm{OBD-SD}},
\label{eq:dml_obdsd}
\end{equation}

\section{Theoretical Support}
\label{sec:theoretical}
In this section, we show why OBD-SD can improve metric learning. We analyze PSD and OBDP, respectively\footnote{We provide the full derivation in Section S.4 of the supplementary.}.

\subsection{Analysis for PSD}
We analyze the gradients of PSD. First, the gradient for $L_{\textrm{PSD}}$ (for convenience, $L_{\textrm{PSD}}$ here is for an anchor $\boldsymbol{x}_{i}$) \emph{w.r.t.} the embedded feature $\boldsymbol{v}_i$ is: 
\begin{equation}
    \frac{\partial L_{PSD}}{\partial \boldsymbol{v}_i} 
        =\sum_{j}^{\left | \mathcal{B} \right |} \left ( \boldsymbol{z}_j -\left ( \boldsymbol{z}_i \cdot \boldsymbol{z}_j \right ) \boldsymbol{z}_i\right ) \left (   P_{ij} - S_{ij}\right ),
\label{eq:support_psd_1}
\end{equation}
where $P_{ij}$ and $S_{ij}$ are the relationships between $\boldsymbol{x}_{i}$ and $\boldsymbol{x}_{j}$ predicted by the student and teacher, respectively. 
We can observe that Eq~\ref{eq:support_psd_1} contains two parts:``attention part" $( \boldsymbol{z}_j \!-\!( \boldsymbol{z}_i \!\cdot\!\boldsymbol{z}_j ) \boldsymbol{z}_i)$ and ``difference part" $\left (  P_{ij}\!-\!S_{ij}\right )$.

First, we analyze the ``attention part". 
Take positive samples as an example, for an easy positive sample $\boldsymbol{z}_{i} \cdot \boldsymbol{z}_{j} \approx 1$, we can achieve:
\begin{equation}
        \|\boldsymbol{z}_{p}-\left(\boldsymbol{z}_{i} \cdot \boldsymbol{z}_{p}\right) \boldsymbol{z}_{i} \|=\sqrt{1-\left(\boldsymbol{z}_{i} \cdot \boldsymbol{z}_{p}\right)^{2}} 
        \approx 0,
\label{eq:support_psd_easy_samples}
\end{equation}
And for an hard positive sample $0 < \boldsymbol{z}_{i} \cdot \boldsymbol{z}_{p} \ll 1$, we have:
\begin{equation}
        \|\boldsymbol{z}_{p}-\left(\boldsymbol{z}_{i} \cdot \boldsymbol{z}_{p}\right) \boldsymbol{z}_{i} \|=\sqrt{1-\left(\boldsymbol{z}_{i} \cdot \boldsymbol{z}_{p}\right)^{2}} 
        \approx 1,
\label{eq:support_psd_hard_samples}
\end{equation}
Obviously, for easy samples, the gradient of $L_{\textrm{PSD}}$ along $\boldsymbol{v}_i$ is close to $0$, while for hard samples is equal to $\|\boldsymbol{z}_{p}-\left(\boldsymbol{z}_{i} \cdot \boldsymbol{z}_{p}\right) \boldsymbol{z}_{i} \|\left|P_{i p}-S_{i p}\right|\approx \left|P_{i p}-S_{i p}\right|$.
This shows that the gradient contribution of easy samples is relatively small while the gradient contribution of hard samples is large. This result implies that PSD encourages the model to pay more attention to hard samples during training. 
We argue that there is usually more intrinsic information (such as intra-class variations) in hard samples that conventional methods would suppress.

Second, we analyze the ``difference part". The ``difference part" shows that PSD forces the student's output $P_{i p}$ to become similar to the teacher's $S_{i p}$, rather than leverage solely class labels to increase inter-class distances and decrease intra-class distances. 
Hence, the student can learn more information in the process of imitation, especially for hard samples, thus alleviating the problem of ignoring the intrinsic relationships of samples.

\subsection{Analysis for OBDP}
Inspired by~\cite{bai2017ensemble,bai2017regularized}, our OBDP can be seen as the following optimization problem:
\begin{equation}
\begin{split}
    &\min _A \frac{1}{2} \!\sum_{i, j, k=1}^{|\mathcal{B}|} \!W_{j,k}\!\left(\frac{A_{i,j}}{\sqrt{V_{i,i} V_{j,j}}}-\frac{A_{i,k}}{\sqrt{V_{i,i} V_{k,k}}}\right)^2\\
    &+\frac{1-\omega}{\omega} \sum_{i,j=1}^{|\mathcal{B}|}\left(A_{i,j}-D_{i,j}\right)^2
\end{split}
\label{eq:support_OBDP}
\end{equation}
where $A$ represents the refined similarity and $D$ is the initial similarity matrix. Eq.~\ref{eq:support_OBDP} consists of two terms. From the first term, we can find that OBDP takes into account high-order relationship information between samples. For example, the relationships between $\boldsymbol{x}_{i}$ and $\boldsymbol{x}_{j}$ (denoted by $A_{i,j}$) is interrelated to relationship between $\boldsymbol{x}_{i}$ and all other samples in the batch (denoted by $A_{i,k}, k=1,\dots,|\mathcal{B}|$) with weights $W_{j,k}$. 
In contrast, $D_{i,j}$ only considers pairwise relationships on rigid distance metrics and ignores the intrinsic relationships between samples.  
We infer that such high-order information on batch manifolds can help the embedding space better reveal the intrinsic relationships between objects as the high-order information on the global manifolds does~\cite{donoser2013diffusion}.
The second term suggests that the initial states should be preserved to a certain extent. It emphasizes that a good distance measurement should not differ too much from the initial states.

\section{Experiments}
\label{sec:experiments}

\subsection{Experimental Details}
\noindent \textbf{Datasets.} To evaluate the generalization on unseen classes, we followed the existing experiment setting from~\cite{roth2020revisiting,roth2021simultaneous}, which is a zero-shot setting where the training set has no intersection with the test set. 
We conduct experiments on three benchmark dataset, such as  \textbf{CUB200-2011}~\cite{wah2011caltech}, \textbf{CARS196}~\cite{krause20133d} and \textbf{Stanford Online Products (SOP)}~\cite{oh2016deep}. 
The CUB200-2011 contains 11,788 images of 200 classes of birds. 
We use the first/last 100 classes (5,864/5,924 images respectively) as the training/test subsets. The CARS196 contains 16,185 images of 196 car categories. We use the first/last 98 classes (8,054/8,131 images) as the training/test subsets. The SOP contains 120,053 online product images in 22,634 categories. We use the first 11,318 classes of 59,551 images as the training subset and the remaining 11,316 classes of 60,502 images as the test subset. 

\begin{table*}[ht]
\tabcolsep=10pt
\caption{Comparison with strong metric learning baselines. \textbf{Bold} denotes the better result between using OBD-SD and not using OBD-SD with the same metric learning loss function.} 
\centering
\resizebox{0.9\linewidth}{!}{
    \begin{tabular}{l|cc|cc|cc}
    \toprule
    Benchmarks$\rightarrow$ & \multicolumn{2}{c}{CUB200-2011} & \multicolumn{2}{c}{CARS196} & \multicolumn{2}{c}{SOP} \\
    \midrule
    Approaches$\downarrow$ & \textbf{R@1}  & \textbf{NMI}  & \textbf{R@1}  & \textbf{NMI}  & \textbf{R@1}  & \textbf{NMI} \\
    \midrule
    Margin ($\beta=1.2$)~\cite{wu2017sampling}              
    & 63.15&  68.05& 80.02& 67.50& 78.20& 90.32\\
    Margin ($\beta=1.2$) + OBD-SD (Ours) 
    &\textbf{65.78($\uparrow$ 2.63)} 
    &\textbf{69.67($\uparrow$ 1.62)} 
    &\textbf{81.49($\uparrow$ 1.47)} 
    &\textbf{68.18($\uparrow$ 0.68)} 
    &\textbf{78.81($\uparrow$ 0.61)}  
    &\textbf{90.51($\uparrow$ 0.19)} \\
    \midrule
    R-Margin ($\beta=0.6$)~\cite{roth2020revisiting}               & 64.54&  68.31& 82.39& 68.57& 77.45& 90.36\\
    R-Margin ($\beta=0.6$) + OBD-SD (Ours) 
    &\textbf{66.54($\uparrow$ 2.00)} 
    &\textbf{69.24($\uparrow$ 0.93)} 
    &\textbf{84.49($\uparrow$ 2.10)} 
    &\textbf{70.24($\uparrow$ 1.67)} 
    &\textbf{77.75($\uparrow$ 0.30)} 
    &\textbf{90.50($\uparrow$ 0.14)}\\
    \midrule
    MS~\cite{wang2019multi}              & 63.10&  68.33& 81.56& 69.38& 77.75& 90.01\\
    MS + OBD-SD (Ours) 
    &\textbf{67.76($\uparrow$ 4.66)} 
    &\textbf{70.81($\uparrow$ 2.48)} 
    &\textbf{84.10($\uparrow$ 2.54)}  
    &\textbf{70.93($\uparrow$ 1.55)}  
    &\textbf{78.83($\uparrow$ 1.08)} 
    &\textbf{90.22($\uparrow$ 0.21)}\\
    
    \bottomrule
    \end{tabular}
    }
\label{tab:results}
\end{table*}
\begin{table*}
\caption{Comparison with state-of-the-art metric learning methods across different Architecture (Arch) and Dimensionality (Dim). ``+\textcolor{red}{D}" indicates that use combinational pooling in backbone as done in~\cite{kim2020proxy}. \textbf{Bold} denotes the best result in the given Arch/Dim setting.  \textcolor{blue}{\textbf{Boldblue}} indicates the best results overall.
}
\centering
\small 
\resizebox{1.0\linewidth}{!}{
    \begin{tabular}{l|l|l||p{1.2cm}<{\centering}p{1.2cm}<{\centering}p{1.0cm}<{\centering}|p{1.2cm}<{\centering}p{1.2cm}<{\centering}p{1.2cm}<{\centering}|p{1.2cm}<{\centering}p{1.2cm}<{\centering}p{1.2cm}<{\centering}}
    \toprule
    \multicolumn{3}{l}{Benchmarks$\rightarrow$} & \multicolumn{3}{c}{CUB200-2011} & \multicolumn{3}{c}{CARS196} & \multicolumn{3}{c}{SOP} \\
    \midrule
    Approaches$\downarrow$ & Venue & Arch/Dim & \textbf{R@1}  & \textbf{R@2}  & \textbf{NMI}  & \textbf{R@1}  &\textbf{R@2}  & \textbf{NMI}  & \textbf{R@1}  &\textbf{R@10}  & \textbf{NMI} \\
    \midrule
    Div\&Conq~\cite{sanakoyeu2019divide}    &CVPR'19    &R50/128    &65.9   &76.6   &69.6  &84.6  &90.7 &70.3  &75.9  &88.4 &90.2\\
    MIC~\cite{roth2019mic}                  &ICCV'19    &R50/128    &66.1   &76.8   &69.7  &82.6  &89.1 &68.4  &77.2  &89.4 &90.0\\
    PADS~\cite{roth2020pads}                &CVPR'20    &R50/128    &67.3   &78.0   &69.9  &83.5  &89.7 &68.8  &76.5  &89.0 &89.9\\
    RankMI~\cite{kemertas2020rankmi}        &CVPR'20    &R50/128    &66.7   &77.2   &\textbf{71.3}  &83.3  &89.8 &69.4  &74.3  &87.9 &90.5\\
    MS+PLG~\cite{roth2022integrating}\tablefootnote{Require pretrained natural language models.} &CVPR'22   &R50/128  &  67.8    &78.2   &  70.1&86.0    &91.4  &72.4  &  77.9   &89.9   & 90.2\\
    \midrule
    \multirow{2}{*}{MS + OBD-SD (Ours)}     &-          &R50/128  &  67.8    &78.1   &  70.8&  84.9 &91.0 & 72.1& \textbf{79.6}  &91.2 & \textbf{90.4}\\
    &-  &R50/128+\textcolor{red}{D}  &  \textbf{68.0}   &\textbf{78.3}    &  71.1   &  \textbf{86.2}&\textbf{91.6}   &  \textbf{73.1}&\textbf{79.6}  &\textbf{91.2} & 90.3\\
    \midrule
    ProxyAnchor (PA)~\cite{kim2020proxy}    &CVPR'20    &IBN/512+\textcolor{red}{D}  &68.4  &79.2 &-  &86.1  &91.6 &-  &79.1  &90.8 &-\\
    ProxyGML~\cite{zhu2020fewer}            &NeurIPS'20 &IBN/512  &66.7 &77.6 &69.8  &85.5 &91.8  &72.4  &78.0 &90.6  &90.4\\
    PA+DRML~\cite{zheng2021deep}            &ICCV'21    &IBN/512  &68.7 &78.6 &69.3  &86.9 &92.1  &72.1  &71.5 &85.2  &88.1\\
    PA+\textit{MemVir}~\cite{ko2021learning}&ICCV'21    &IBN/512  &69.0 &79.2 &-  &86.7 &92.0  &-  &79.7 &91.0  &-\\
    EPSHN~\cite{xuan2020improved}           &WACV'20    &R50/512  &64.9 &75.3 &-  &82.7 &89.3  &-  &78.3 &90.7  &-\\
    DiVA~\cite{milbich2020diva}             &ECCV'20    &R50/512  &69.2 &79.3 &71.4  &87.6 &92.9  &72.2  &79.6 &91.2  &90.6\\
    DCML-MDW~\cite{zheng2021deepcompos}     &CVPR'21    &R50/512  &68.4 &77.9 &\textcolor{blue}{\textbf{71.8}}  &85.2 &91.8  &73.9  &79.8 &90.8  &\textcolor{blue}{\textbf{90.8}}\\
    MS+DAS~\cite{liu2022densely}            &ECCV'22    &R50/512+\textcolor{red}{D}  &69.2  & 79.3&-  &87.8 &93.1  &-  &80.6 &91.8  &-\\
    \midrule
    \multirow{4}{*}{MS + OBD-SD (Ours)}    &-  &IBN/512  &  65.8&76.8 & 68.8& 85.4 &91.4  &70.4  &77.6 &90.0  &89.6\\
    &-  &IBN/512+\textcolor{red}{D}  &65.8 &76.9 &68.7 &85.8 &92.1 &  71.6 &77.9 &90.1  &89.5\\
    &-  &R50/512  &  69.1&79.5 &  71.1 &86.5 &92.3  &72.0  &80.7 &92.0 & 90.4\\
    &-  &R50/512+\textcolor{red}{D}  &  \textcolor{blue}{\textbf{69.4}}&\textcolor{blue}{\textbf{79.7}} &  71.4 &\textcolor{blue}{\textbf{88.7}} & \textcolor{blue}{\textbf{93.6}}  &\textcolor{blue}{\textbf{74.5}}  &\textcolor{blue}{\textbf{81.0}} &\textcolor{blue}{\textbf{92.1}}  &90.1\\
    \bottomrule
\end{tabular}
    }
\label{tab:results_sota}
\end{table*}

\noindent \textbf{Evaluation Metrics.} Recall@K (R@K)~\cite{jegou2010product} and Normalized Mutual Information (NMI)~\cite{manning2008introduction} are used, where R@k measures the image retrieval performance while NMI measures the image clustering performance. Please see the supplementary (Section S.1) for more information.

\noindent \textbf{Implementation Details.} We utilize a ImageNet pretrained ResNet50 (R50/d) or Inception BN (IBN/d) with frozen Batch-Normalization layers as the backbone network, where the $d$ is the embedding dimension. If not specifically emphasized, we use R50/128 by default in experiments. 
We use Adam optimizer with a fixed learning rate of $10^{-5}$ and a weight decay of $4 \times 10^{-4}$. 
Note that there is no learning rate scheduling for unbiased comparison, if not mentioned.
We set the distillation weight $\lambda$/transition probability $\omega$ to 1000/0.3 for CUB200-2011, 75/0.99 for CARS196 and 100/0.5 for Stanford Online Products, and the temperature to 1 by default. Each training was run over $150$ epochs for CUB200-2011/CARS196 and $100$ epochs for Stanford Online Products. All experiments were performed on a single Nvidia V100 GPU and all results are computed over multiple seeds averages. Additional details are available in the supplementary (Section S.3).

\subsection{Effectiveness of OBD-SD}
To validate the effectiveness, we use R50/128 as the backbone. 
We select three methods that can achieve top-3 performance evaluated in~\cite{roth2020revisiting}, \textit{i.e.}, Multisimilarity (MS) loss~\cite{wang2019multi}, Margin loss with Distance-based Sampling~\cite{wu2017sampling} and Regularized Margin loss~\cite{roth2020revisiting}, and follow their experimental training pipeline. 
Note that the setup utilizes no learning rate scheduling and fixes common implementational factors of variation (such as batch-size) in DML pipelines to ensure comparability. 
The results are presented in Table~\ref{tab:results}, which shows that our OBD-SD improves the original DML approaches on all benchmarks.

To further highlight the benefits, we compare our OBD-SD with other state-of-the-art methods and report results in Table~\ref{tab:results_sota}. The compared methods can be divided into different groups (\emph{i.e.,} R50/128, R50/512 and IBN/512), according to their backbone and embedding dimensions. 
Note that we apply learning rate scheduling and extend the number of the training epoch appropriately. 
The results in Table~\ref{tab:results} show that OBD-SD can achieve competitive performance across different architectures and benchmarks. 

\subsection{Ablation Studies}
\label{sec:abla}
\noindent \textbf{Ablation study of each element.} We perform ablation studies to evaluate the effectiveness of each element in OBD-SD. 
We use MS loss as the default DML loss.
The results are shown in Table~\ref{tab:abla_elements}. 
We observe that our OBD-SD (PSD with OBDP) achieves the best result, demonstrating the effectiveness of each proposed element. 
For the ``+PSD w/o dynamic weight", we conduct ablation experiments to select the best $\lambda$ value.
\begin{table}
  \caption{Ablation study using different elements of OBD-SD.}
  \tabcolsep=10pt
  \centering
  \resizebox{1.0\linewidth}{!}{
  \begin{tabular}{lcccc}
    \toprule
    Datasets$\rightarrow$ & \multicolumn{2}{c}{CUB200-2011} & \multicolumn{2}{c}{CARS196} \\
    \midrule
    Methods$\downarrow$    &\textbf{R@1}    &\textbf{NMI}    &\textbf{R@1}    &\textbf{NMI} \\
    \midrule
    MS~\cite{wang2019multi} & 63.1&  68.3&   81.6&   69.4\\
    + PSD w/o dynamic weight & 62.9&   68.6&   81.9&   69.4\\
    + PSD & 63.5&   69.1&   82.3&   70.0\\
    + OBD-SD (PSD+OBDP)   & \textbf{67.8}&   \textbf{70.8}&   \textbf{84.1}&   \textbf{70.9}\\
    \bottomrule
  \end{tabular}
  }
  \label{tab:abla_elements}
\end{table}

\noindent \textbf{Ablation Studies on Different Sampling Strategies.} 
Data sampling is an important technique for pair-based DML methods. To analyze the impact of data sampling on OBD-SD, we employ four popular data sampling strategies: 1) random sampling, 2) semi-hard sampling~\cite{schroff2015facenet}, 3) soft-hard mining~\cite{roth2019mic}, and 4) distance-based tuple mining~\cite{wu2017sampling} with Margin loss~\cite{wu2017sampling} for the experiments. 
We report the results in Table~\ref{tab:sampling}. We find that OBD-SD can consistently improve performance through different sampling strategies. 
\begin{table}
  \caption{Ablation studies on different sampling strategies. The suffixes `(R)', `(H)', `(S)' and `(D)' denote the random, semi-hard, soft-hard and distance-based sampling strategy, respectively. }
  \tabcolsep=15pt
  \centering
  \resizebox{1.0\linewidth}{!}{
  \begin{tabular}{lcccc}
    \toprule
    Datasets$\rightarrow$ & \multicolumn{2}{c}{CUB200-2011} & \multicolumn{2}{c}{CARS196} \\
    \midrule
    Methods$\downarrow$    &\textbf{R@1}    &\textbf{NMI}    &\textbf{R@1}    &\textbf{NMI} \\
    \midrule
    Margin (R) & 55.5&  61.7&  66.1&  59.5\\
    + OBD-SD & \textbf{62.3}&   \textbf{67.1}&   \textbf{73.3}&   \textbf{63.1}\\
    \midrule
    Margin (H)~\cite{schroff2015facenet} & 61.2&  66.5&  76.1&  64.4\\
    + OBD-SD & \textbf{63.6}&   \textbf{67.8}&   \textbf{78.3}&   \textbf{65.6}\\
    \midrule
    Margin (S)~\cite{roth2019mic} & 61.3&  66.5&  77.7&  66.1\\
    + OBD-SD & \textbf{63.3}&   \textbf{68.1}&  \textbf{78.6}&   \textbf{66.4}\\
    \midrule
    Margin (D)~\cite{wu2017sampling} & 63.2&  68.1&  80.0&  67.5\\
    + OBD-SD & \textbf{65.8}&   \textbf{69.7}&   \textbf{81.5}&   \textbf{68.2}\\
    
    \bottomrule
  \end{tabular}
  }
  \label{tab:sampling}
\end{table}

\subsection{Comparison with Self-KD of DML.}
\label{subsec:comparision_self-kd}
Then, we compare OBD-SD with recently proposed Self-KD-based DML methods, such as BAR~\cite{qin2021born} and S2SD~\cite{roth2021simultaneous}.
We use MS loss as the basic DML loss for all experiments in this part. 
For BAR, we select the best distillation weights by ablation experiments (100/50 for CUB200/CARS196) since the paper doesn't show the experiments on the datasets we used, and for S2SD, we follow the setting of its paper.
The results are shown in Table~\ref{tab:comparison_self_kd}. Compared with BAR, our proposed Self-KD method, PSD, can achieve competitive results with only $5.3\%$ and $1.9\%$ extra training time on CUB200 and CARS196 datasets, while BAR requires almost double training time. 
We also compare OBD-SD with S2SD, a SOTA Self-KD method for DML. 
The results in Table~\ref{tab:comparison_self_kd} show that our OBD-SD requires much less extra training time, $17.4\%$ and $7.4\%$ on the two datasets, and achieves competitive performance. 

Since our OBDP is a plug-and-play technique, we combine OBDP with recent Self-KD-based DML, \emph{i.e.,} BAR and S2SD, to study the benefits of OBDP.
Specifically, similar to our OBD-SD, we use OBDP to refine the soft labels generated by teachers (in BAR and S2SD), then use the refined soft labels for knowledge distillation. 
The results are reported in Table~\ref{tab:self_kd_obdp}, which shows that OBDP can also improve the performance of other Self-KD methods.

\begin{table}
  \caption{Comparison with other self-distillation methods of DML.}
  \tabcolsep=6pt
  \centering
  \resizebox{1.0\linewidth}{!}{
  \begin{tabular}{lcccccc}
    \toprule
    Datasets$\rightarrow$ & \multicolumn{3}{c}{CUB200-2011} & \multicolumn{3}{c}{CARS196} \\
    \midrule
    Methods$\downarrow$    &R@1   &NMI    &GPU-Time    &R@1    &NMI    &GPU-Time\\
    \midrule
    MS~\cite{wang2019multi} & 63.1&   68.3&   -&   81.6& 69.4& -\\
    \midrule
    MS+BAR~\cite{qin2021born} & 64.1&   68.4&   +105.3\%&   82.6& 69.6& +101.9\%\\
    MS+PSD (Ours) & 63.5&   69.1&   +5.3\%&   82.3& 70.0&+1.9\%\\
    \midrule
    MS+S2SD-MSD~\cite{roth2021simultaneous}\tablefootnote{We carefully reproduce the experiments by using the official code.} & 65.2&   70.6&   +50.4\%&   84.0& 71.3&+35.4\%\\
    MS+OBD-SD (Ours)& 67.8&   70.8&   +17.4\%&   84.1& 70.9&+7.4\%\\
    \bottomrule
  \end{tabular}
  }

  \label{tab:comparison_self_kd}
\end{table}

\begin{table}
  \caption{Experimental results of applying OBDP to other self-distillation of methods of DML.}
  \tabcolsep=10pt
  \centering
  \resizebox{1.0\linewidth}{!}{
  \begin{tabular}{lcccc}
    \toprule
    Datasets$\rightarrow$ & \multicolumn{2}{c}{CUB200-2011} & \multicolumn{2}{c}{CARS196} \\
    \midrule
    Methods$\downarrow$&\textbf{R@1}    &\textbf{NMI}    &\textbf{R@1}    &\textbf{NMI} \\
    \midrule
    MS+BAR~\cite{qin2021born} & 64.1&   68.4&   82.6&   69.6\\
    MS+BAR+OBDP & \textbf{66.7}&   \textbf{70.2}&   \textbf{84.5}&   \textbf{71.4}\\
    \midrule
    MS+S2SD-MSD~\cite{roth2021simultaneous} & 65.2&   70.6&   84.0&   71.3\\
    MS+S2SD-MSD+OBDP &\textbf{65.8}  &\textbf{70.9}   &\textbf{84.6}   &\textbf{71.9}   \\
    \bottomrule
  \end{tabular}
  }
  \label{tab:self_kd_obdp}
\end{table}

\subsection{Batch Manifolds versus Global Manifolds}

In this section, we conduct experiments comparing the diffusion in batch manifolds and in global manifolds. 
We use batch diffusion and global diffusion (Section~\ref{subsec:preliminaries}) for refining the soft distance targets, respectively.
As mentioned in Section~\ref{sec:OBDP}, Eq.~\ref{eq:diffusion} requires $O(N^3)$ time complexity, so it's impossible for us to update it for each mini-batch (online diffusion). 
Instead, in our experiments, we update the matrix once after each training epoch (offline diffusion). 
Specifically, after the training of each epoch, we use the new teacher to extract features of the training set and calculate the refined similarity matrix for all data. 
When we train the student in the next epoch, for each batch, we select the related elements from the refined similarity matrix for knowledge distillation. 
Note that this strategy leads to ignoring some variation factors, such as those caused by data augmentations. 
In addition, to reduce the space complexity and the noise from negative pairs, we construct the affinity matrix $W$ as follows:
\begin{equation}
\small
    W_{i,j}\!=\!\begin{cases}d\left(\boldsymbol{z}_{i}, \boldsymbol{z}_{j}\right)\!\!& \!\!i \neq j, \boldsymbol{x}_{i}\!\in\! \mathrm{NN}_k\!\left(\boldsymbol{x}_{j}\!\right), \boldsymbol{x}_{j}\!\in\! \mathrm{NN}_k\left(\boldsymbol{x}_{i}\right) \\ 0 & \text { otherwise }\end{cases}
\end{equation}
where $\mathrm{NN}_k\left(\boldsymbol{z}\right)$ denotes the top-k nearest samples of $\boldsymbol{x}$. 
This way can also improve the speed of diffusion. 
In our experiments, we set $k = 50$ and use MS loss, the size of the training set for CUB200 and CARS196 datasets are $5,864$ and $8,054$ respectively, and the size of the mini-batch is $112$. The results are reported in Table~\ref{tab:batch_vs_global}. From the results, we can find that diffusion in batch manifold can achieve almost the same result as diffusion in global manifold and with much less extra time. 

\begin{table}
  \caption{The results of ``Batch Manifolds vs Global Manifolds".}
  \tabcolsep=2pt
  \centering
  \resizebox{1.0\linewidth}{!}{
  \begin{tabular}{lcccccc}
    \toprule
    Datasets$\rightarrow$ & \multicolumn{3}{c}{CUB200-2011} & \multicolumn{3}{c}{CARS196} \\
    \midrule
    Methods$\downarrow$&\textbf{R@1}    &\textbf{NMI} &\textbf{GPU-Time} &\textbf{R@1}    &\textbf{NMI} &\textbf{GPU-Time} \\
    \midrule
    In global manifold & 67.3&   70.6&   +248.8\%&   84.4&  71.0& +345.0\%\\
    In batch manifold (Ours) & 67.8&70.8 &+17.4\%   &84.1   &70.9   &+7.4\% \\
    \bottomrule
  \end{tabular}
  }
  \label{tab:batch_vs_global}
\end{table}

\subsection{Embedding Space Metrics}
\label{subsection:space_metrics}
We investigate the effect of OBD-SD on embedding space. 
We observe the differences between the embedding space learned with and without OBD-SD by using different structural measurements. (1) \textit{Embedding Space Density}~\cite{roth2020revisiting}: $\pi_{\text {ratio }}(\Phi)=\pi_{\text {intra }}(\Phi) / \pi_{\text {inter }}(\Phi)$, which measures the ratio of mean inter-class distances $\pi_{\textit{inter}}\left(\Phi\right)$ and mean intra-class distances $\pi_{\text {intra }}(\Phi)$. 
Note that the higher value of embedding space density indicates lower class concentration and is linked to stronger generalization. 
(2) \textit{Spectral Decay}~\cite{roth2020revisiting}: $\rho(\Phi) = \operatorname{KL}(\mathcal{U} \| S(\Phi))$, where $\rho(\Phi)$ calculates the KL-divergence between uniform distribution $\mathcal{U}$ and singular value decomposition $S(\Phi)$ of embedding space $\Phi$. 
The spectral decay measures the number of directions of significant variance in the learned embedding space. 
Note that the lower value of spectral decay indicates high feature variety and is linked to stronger generalization. In our experiments, we exclude the two largest spectral values for a more robust estimate. Please see the supplementary (Section S.2) for more information.

The experimental results are reported in Table~\ref{tab:space_metrics}. 
The results show that both PSD and OBDP improve the embedding space density and reduce the spectral decay, linked to more robust generalization as described above. 
In other words, OBD-SD can reduce the over-clustering in the embedding space and increase embedding representation diversity, which means OBD-SD can encourage the model to find more intrinsic relation information. The result is consistent with our initial motivation for OBD-SD.

\begin{table}
  \caption{Results of embedding space density $\pi_{\text {ratio }}(\Phi)$ and spectral decay $\rho(\Phi)$.}
  \tabcolsep=8pt
  \centering
  \resizebox{1.0\linewidth}{!}{
  \begin{tabular}{lcccc}
    \toprule
    Datasets$\rightarrow$ & \multicolumn{2}{c}{CUB200-2011} & \multicolumn{2}{c}{CARS196} \\
    \midrule
    Methods$\downarrow$&\textbf{$\pi_{\text {ratio }}(\Phi)$}    &\textbf{$\rho(\Phi)$}    &\textbf{$\pi_{\text {ratio }}(\Phi)$}    &\textbf{$\rho(\Phi)$} \\
    \midrule
    MS & 0.4427&   0.1544&   0.3962&   0.1622\\
    MS+PSD & 0.4504$\uparrow$&  0.1444$\downarrow$ &0.4256$\uparrow$  &0.1496$\downarrow$ \\
    MS+OBD-SD & \textbf{0.6291$\uparrow$}&   \textbf{0.0292$\downarrow$}&   \textbf{0.7083$\uparrow$}&   \textbf{0.0264$\downarrow$}\\
    \bottomrule
  \end{tabular}
  }
  \label{tab:space_metrics}
\end{table}

\subsection{Robustness to Mislabeled Data}
Since our approach alleviates the constraints imposed by sample annotation, a natural question is raised: \textit{Does it also improve the robustness to mislabeled data? }

To answer this question, we generate mislabeled data on the three datasets (CUB200, CARS196 and Stanford Online Products) with $\{0.1, 0.2, 0.3, 0.4\}$ symmetric mislabeled ratios, following \cite{han2018co} and \cite{van2015learning}. 
We then use MS loss as the baseline and train models on the mislabeled data with and without OBD-SD. 
The results in Table~\ref{tab:noisy} indicate that OBD-SD consistently achieves performance promotions on mislabeled training data. 
Moreover, we calculate the percentage of improvement relative to the baseline for each metric as follows: $P_{metric}=(V_{SD}-V_{base})/V_{base}$, where $V_{SD}$ and $V_{base}$ are the value of the metric for OBD-SD and the baseline, respectively. We can find that the performance promotions become more pronounced as the mislabeled ratio increases. 
This result shows that OBD-SD can improve the robustness to mislabeled data.
\tabcolsep=4pt
\begin{table}
\caption{Results on the three datasets validation with different noisy ratios on the training sets.}
\centering
\resizebox{1.0\linewidth}{!}{
    \begin{tabular}{cccccc}
    \toprule
    \multirow{2}{*}{Noisy ratio} & Benchmarks$\rightarrow$ & \multicolumn{2}{c}{CUB200} & \multicolumn{2}{c}{CARS196}\\
    &Approaches$\downarrow$ & \textbf{R@1} & \textbf{NMI} & \textbf{R@1} &  \textbf{NMI}\\
    \midrule
    \multirow{3}{*}{0.1} 
    & MS        &59.69   &66.70   &76.81   &65.26   \\
    & \multirow{2}{*}{MS+OBD-SD} 
    &\textbf{66.11}   &\textbf{69.93}   
    &\textbf{80.47}   &\textbf{67.85}   \\
    &&($\mathbf{+10.75\%}$)   &($\mathbf{+4.84\%}$)   &($\mathbf{+4.77\%}$)   &($\mathbf{+3.97\%}$)  \\
    \midrule
    \multirow{2}{*}{0.2} 
    & MS        &55.83   &63.92   &71.33   &60.92   \\
    & \multirow{2}{*}{MS+OBD-SD}
    &\textbf{64.23}   &\textbf{68.84}   
    &\textbf{76.57}   &\textbf{65.09}   \\
    &&($\mathbf{+15.05\%}$)   &($\mathbf{+7.70\%}$)   &($\mathbf{+7.35\%}$)   &($\mathbf{+6.85\%}$)  \\
    \midrule
    \multirow{2}{*}{0.3} 
    & MS        &52.03   &61.06   &66.53   &56.24   \\
    & \multirow{2}{*}{MS+OBD-SD}
    &\textbf{61.83}   &\textbf{67.00}   
    &\textbf{71.16}   &\textbf{59.98}   \\
    &&($\mathbf{+18.84\%}$)   &($\mathbf{+9.73\%}$)   &($\mathbf{+6.96\%}$)   &($\mathbf{+6.65\%}$)  \\
    \midrule
    \multirow{2}{*}{0.4} 
    & MS        &48.25   &57.75   &60.97   &51.16   \\
    & \multirow{2}{*}{MS+OBD-SD}
    &\textbf{58.11}   &\textbf{65.05}   
    &\textbf{65.58}   &\textbf{54.43}   \\
    &&($\mathbf{+20.44\%}$)   &($\mathbf{+13.07\%}$)   &($\mathbf{+7.56\%}$)   &($\mathbf{+6.39\%}$)  \\
    \midrule

    \bottomrule
    \end{tabular}
}
\label{tab:noisy}
\end{table}
\section{Conclusion}
In this paper, we propose Online Batch Diffusion-based Self-distillation (OBD-SD) to alleviate the "ignore intrinsic relation" issue of DML methods. 
To this end, we propose Progressive Self-distillation (PSD) to encourage the model to learn more relational information among samples. Then, we further propose Online Batch Diffusion Process (OBDP) to help PSD reveal the intrinsic relationships among samples on batch manifolds.
Extensive experiments with various loss functions on three benchmarks show that OBD-SD is effective and efficient. 
However, our proposed method can't consistently improve the performance of proxy-based DML, which is a strong method of DML. In the feature, we plan to explore the diffusion process on the proxies-manifold for optimizing the embedding space for proxy-based DML. 
In addition, we also plan to apply OBD-SD to other areas such as self-supervised learning



\section{Supplemental Material}
You can download the Supplemental Material from \small{\url{https://github.com/ZelongZeng/OBD-SD_Pytorch}.}

\section{Acknowledgements}

This research is part of the results of Value Exchange Engineering, a joint research project between Mercari, Inc. and the RIISE of the University of Tokyo.

{\small
\bibliographystyle{ieee_fullname}
\bibliography{Main}
}
\end{document}


\title{The Supplementary of ``Self-distillation with Online Diffusion on Batch Manifolds Improves Deep Metric Learning''}

\author{Zelong Zeng\\
The University of Tokyo\\
\& National Institute of Informatics\\
{\tt\small zzlbz@nii.ac.jp}
\and
Fan Yang\\
National Institute of Informatics\\
{\tt\small yang@nii.ac.jp}
\and
Hong Liu\\
National Institute of Informatics\\
{\tt\small hliu@nii.ac.jp}
\and
Shin'ichi Satoh\\
National Institute of Informatics\\
\& The University of Tokyo\\
{\tt\small satoh@nii.ac.jp}
}
\maketitle

\begin{abstract}
The following items are included in the supplementary material.
\begin{itemize}
\item Evaluation Metrics Details
\item Embedding Space Metrics Details
\item More Experimental Implementation Details
\item Propositions.
\item Ablation Studies on Hyper-parameters.
\item Visualization.
\item Code.
\end{itemize}
\end{abstract}

\section{Evaluation Metrics Details}
The evaluation metrics used in this work are Recall@K (R@K)~\cite{jegou2010product} and Normalized Mutual Information (NMI)~\cite{manning2008introduction}.
\noindent \textbf{Recall@K} is the primary metric used to compare the performance of DML methods. Let $\mathcal{X}_{\text {test }}$ as the test subset and $\mathcal{F}_q^k$ is a set of the first k nearest neighbors of a sample $x_q$, then we measure Recall@k as
\begin{equation}
    R @ k=\frac{1}{\left|\mathcal{X}_{\text {test }}\right|} \sum_{x_q \in \mathcal{X}_{\text {test }}} \begin{cases}1 & \exists x_i \in \mathcal{F}_q^k \text { s.t. } y_i=y_q \\ 0 & \text { otherwise }\end{cases}
\end{equation}
which measures the average number of cases for a given query $x_q$ with at least one sample $x_i$ with the same class among its top k nearest neighbors.

\noindent \textbf{Normalized Mutual Information (NMI)} is used to measure the clustering quality. We first embed and normalize all samples $x_i\in \mathcal{X}_{\text {test }}$ ($\mathcal{X}_{\text {test }}$ indicates the test subset). Then, we use a clustering method (such as K-Means) to calculate the $K=|C|$ clusters where $|C|$ is the number of classes. After that, we can assign each sample $x_i$ a cluster label $w_i$ indicating the closest cluster center. Define $\Omega=\left\{\omega_k\right\}_{k=1}^K$ with $\omega_k=\left\{i \mid w_i=k\right\}$ and $\Upsilon=\left\{v_c\right\}_{c=1}^K$ with $v_c=\left\{i \mid y_i=c\right\}$ ($y_i$ is the true labels), the normalized mutual information can be computed as
\begin{equation}
    N M I(\Omega, \Upsilon)=\frac{I(\Omega, \Upsilon)}{2(H(\Omega)+H(\Upsilon)}
\end{equation}
with mutual Information $I(\cdot,\cdot)$ between cluster and labels, and entropy $H(\cdot,\cdot)$ on the clusters and labels respectively.

\section{Embedding Space Metrics Details}
The embedding space metrics used in this work are Embedding Space Density~\cite{roth2020revisiting} and Spectral Decay~\cite{roth2020revisiting}. We use them to analyze the embedding space of the training subset $\mathcal{X}$.

\noindent \textbf{Embedding Space Density.} we define the embedding space density as $\pi_{\text {ratio }}(\Phi)=\pi_{\text {intra }}(\Phi) / \pi_{\text {inter }}(\Phi)$, where average inter-class distances $\pi_{\textit{inter}}\left(\Phi\right)$ is,

\begin{equation}
\small
    \pi_{\text{inter}}\left(\Phi\right) = \frac{1}{Z_{\text {inter }}} \sum_{\substack{y_{l}, y_{k} \\ l \neq k}} d\left(\mu\left(\Phi_{y_{l}}\right), \mu\left(\Phi_{y_{k}}\right)\right)
\label{eq:inter}
\end{equation}
and average intra-class distances $\pi_{\text {intra }}(\Phi)$ is,

\begin{equation}
\small
    \pi_{\text {intra }}(\Phi)=\frac{1}{Z_{\text {intra }}} \sum_{y_{l} \in \mathcal{Y}} \sum_{\substack{\boldsymbol{z}_{i}, \boldsymbol{z}_{j} \in \Phi_{y_{l}}\\ i \neq j}} d\left(\boldsymbol{z}_{i}, \boldsymbol{z}_{j}\right)
\label{eq:intra}
\end{equation}
Here, $\Phi_{y_{l}}=\left\{\boldsymbol{z}_{i}:=\left \|f\left(\boldsymbol{x}_{i},\theta\right) \right \|_{2}\mid \boldsymbol{x}_{i} \in\mathcal{X}, y_{i}=y_{l}\right\}$ denotes the set of embedded samples of a class $y_{l}$, $\mu\left(\Phi_{y_{l}}\right)$ denotes their mean embedding and $Z_{\text {intra }}$ and $Z_{\text {inter }}$ are normalization constants. Some work~\cite{roth2020revisiting,roth2021simultaneous} have shown that optimizing the DML problem while maintaining a high embedding spaces density, \textit{i.e.}, without over-clustering, can encourage better generalization to unseen test classes.

\noindent \textbf{Spectral Decay.} The spectral decay metric $\rho(\Phi)$ defines the KL-divergence between the (sorted) spectrum of $D$ singular values $S(\Phi)$ (obtained via Singular Value Decomposition (SVD)) and a $D$-dimensional discrete uniform distribution $\mathcal{U}$:
\begin{equation}
    \rho(\Phi)=\operatorname{KL}(\mathcal{U} \| S(\Phi))
\end{equation}
which is inversely related to the entropy of the embedding space. Some works~\cite{roth2020revisiting,roth2021simultaneous} show that optimizing the DML problem while encouraging a high spectral decay, \textit{i.e.}, high feature diversity, notably benefits the generalization performance for DML.

\section{More Experimental Implementation Details}
In this section, we provide more implementation details. As for data argumentation, we randomly resize and crop images to $224 \times 224$ for training, and center crops them to the same size at evaluation. Random horizontal flipping ($p=0.5$) is also utilized in training. In terms of mini-batch, we set batchsize as 112, and randomly select 2 different images for each class.

For the experiments about comparison with other state-of-the-art methods (in Table \textcolor{red}{2} of the main paper), we extend the number of training epoch and use learning rate scheduling for CARS196 and SOP datasets. Specifically, the number of training epoch is extended to 300 and 200 for CARS196 and SOP datasets. Please see the ``SampleRuns.sh" in our uploading code for more information.

\section{Propositions}
\label{sec:prop}
\noindent \textbf{Proposition 1.} The gradient for Eq \textcolor{red}{1} with respect to the embedding $\boldsymbol{v}_i$ is Eq \textcolor{red}{8}. 

\noindent \textit{Proof.} For each anchor $\boldsymbol{x}_i$, PSD function can be described as:
\begin{equation}
\small
    L_{\textup{PSD}}(\mathcal{B}) = 
    \displaystyle\sum_{j}^{\left | \mathcal{B} \right |}\sigma(D_{i,:}^{(T)}/\tau)_j\textup{ln}\frac{\sigma(D_{i,:}^{(T)}/\tau)_j}{\sigma(D_{i,:}^{(S)}/\tau)_j},
\label{appeq:lsd}
\end{equation}
For convenience, let's define:
\begin{equation}
\small
    \sigma(D_{i,:}^{(T)}/\tau)_j = S_{ij},
\end{equation}
\begin{equation}
\small
    D_{i,:}^{(S)} = \boldsymbol{z}_i \cdot \boldsymbol{z}_j,
\end{equation}
\begin{equation}
\small
    B_{PSD} = 
    \displaystyle\sum_{j}^{\left | \mathcal{B} \right |}\sigma(D_{i,:}^{(T)}/\tau)_j\textup{ln}\sigma(D_{i,:}^{(T)}/\tau)_j,
\end{equation}
Then, the PSD function can be described as:
\begin{equation}
\small
    \begin{split}
        L_{PSD} = -\displaystyle\sum_{j}^{\left | \mathcal{B} \right |}S_{ij}\textup{ln}\frac{\textup{exp}(\boldsymbol{z}_i \cdot \boldsymbol{z}_j/\tau)}{\sum_{k}^{\left | \mathcal{B} \right |}\textup{exp}(\boldsymbol{z}_i \cdot \boldsymbol{z}_k/\tau)} + B_{PSD},
    \end{split}
\label{appeq:lsd_convi}
\end{equation}
We now divide the gradient of Eq~\ref{appeq:lsd_convi} with respect to $\boldsymbol{v}_i$:
\begin{equation}
\small
    \frac{\partial L_{PSD}}{\partial \boldsymbol{v}_i} = \frac{\partial L_{PSD}}{\partial \boldsymbol{z}_i} \frac{\partial \boldsymbol{z}_i}{\partial \boldsymbol{v}_i},
\end{equation}
For the $\frac{\partial \boldsymbol{z}_i}{\partial \boldsymbol{v}_i}$, since $\boldsymbol{z}_i = \frac{\boldsymbol{v}_i}{\left \| \boldsymbol{v}_i \right \|}$, we can deduce that:
\begin{equation}
\small
    \begin{split}
        \frac{\partial \boldsymbol{z}_i}{\partial \boldsymbol{v}_i}
        &= \frac{\partial}{\partial \boldsymbol{v}_i} \left ( \frac{\boldsymbol{v}_i}{\left \| \boldsymbol{v}_i \right \|} \right )
        \\
        &=\frac{1}{\left \| \boldsymbol{v}_i \right \|} \textbf{I} - \frac{ \boldsymbol{v}_i \cdot \boldsymbol{v}_i  }{\left \| \boldsymbol{v}_i \right \|}
        \\
        &=\frac{1}{\left \| \boldsymbol{v}_i \right \|} \left (\textbf{I} - \frac{ \boldsymbol{v}_i \cdot \boldsymbol{v}_i  }{\left \| \boldsymbol{v}_i \right \|^2}\right )
        \\
        &=\frac{1}{\left \| \boldsymbol{v}_i \right \|} \left (\textbf{I} - \boldsymbol{z}_i \cdot \boldsymbol{z}_i\right ),
    \end{split}
    \label{appeq:lsd_grad_z2v}
\end{equation}

For the gradient with respect to $\boldsymbol{z}_i$:
\begin{equation}
    \small
    \begin{split}
        &\frac{\partial L_{PSD}}{\partial \boldsymbol{z}_i}
        \\
        &=-\sum_{j}^{\left | \mathcal{B} \right |}\frac{\partial}{\partial \boldsymbol{z}_i}S_{ij} \left \{ (\frac{\boldsymbol{z}_i \cdot \boldsymbol{z}_j}{\tau})-\textup{ln} \sum_{k}^{\left | \mathcal{B} \right |}\textup{exp}\left( \frac{\boldsymbol{z}_i \cdot \boldsymbol{z}_k}{\tau} \right) \right \}
        \\
        &=-\sum_{j}^{\left | \mathcal{B} \right |}\left \{ S_{ij} \left [ \boldsymbol{w}_{j} \boldsymbol{z}_j -\frac{\sum_{k}^{\left | \mathcal{B} \right |} \boldsymbol{w}_{k}\boldsymbol{z}_k \textup{exp} \left( \frac{\boldsymbol{z}_i \cdot \boldsymbol{z}_k}{\tau} \right)}{\sum_{k}^{\left | \mathcal{B} \right |}\textup{exp} \left( \frac{\boldsymbol{z}_i \cdot \boldsymbol{z}_k}{\tau} \right)}  \right ] \right \}
        \\
        &=-\left \{\sum_{j}^{\left | \mathcal{B} \right |} S_{ij} \boldsymbol{w}_{j}\boldsymbol{z}_j - \sum_{j}^{\left | \mathcal{B} \right |}S_{ij} \sum_{k}^{\left | \mathcal{B} \right |} \boldsymbol{w}_{k}\boldsymbol{z}_k P_{ik}
        \right \}
        \\
        &=-\left \{\sum_{j}^{\left | \mathcal{B} \right |} S_{ij} \boldsymbol{w}_{j}\boldsymbol{z}_j - \sum_{k}^{\left | \mathcal{B} \right |} P_{ik} \boldsymbol{w}_{k}\boldsymbol{z}_k 
        \right \}
        \\
        &=-\sum_{j}^{\left | \mathcal{B} \right |} \boldsymbol{w}_{j}\boldsymbol{z}_j \left (  S_{ij}-P_{ij}
        \right ) 
    \end{split}
\label{appeq:lsd_grad_l2z}
\end{equation}
where:
\begin{equation}
\small
    \begin{split}
        P_{ik} = \frac{ \textup{exp} \left( \frac{\boldsymbol{z}_i \cdot \boldsymbol{z}_k}{\tau} \right)}{\sum_{k}^{\left | \mathcal{B} \right |}\textup{exp} \left( \frac{\boldsymbol{z}_i \cdot \boldsymbol{z}_k}{\tau} \right)} 
    \end{split}
\end{equation}
and
\begin{equation}
\small
    \boldsymbol{w}_{j}=\begin{cases}
        1 & \text{ if } j \neq i \\ 
        2 & \text{ if } j = i,
    \end{cases}
\end{equation}
Combining Eq~\ref{appeq:lsd_grad_z2v} and Eq~\ref{appeq:lsd_grad_l2z} thus gives:
\begin{equation}
\small
    \begin{split}
        \frac{\partial L_{PSD}}{\partial \boldsymbol{v}_i} 
        &= -\left (\textbf{I} - \boldsymbol{z}_i \cdot \boldsymbol{z}_i\right ) \sum_{j}^{\left | \mathcal{B} \right |} \boldsymbol{w}_{j}\boldsymbol{z}_j \left (  S_{ij}-P_{ij}\right ) \\
        &=\sum_{j}^{\left | \mathcal{B} \right |} w_j\left ( \boldsymbol{z}_j -\left ( \boldsymbol{z}_i \cdot \boldsymbol{z}_j \right ) \boldsymbol{z}_i\right ) \left (   P_{ij} - S_{ij}\right )
    \end{split}
    \label{appeq:final_grad}
\end{equation}
When $w_j=2$ (\textit{i.e.},when $i=j$), $\left ( \boldsymbol{z}_j -\left ( \boldsymbol{z}_i \cdot \boldsymbol{z}_j \right ) \boldsymbol{z}_i\right )=0$. So, for convenience, Eq~\ref{appeq:final_grad} can be described as:
\begin{equation}
\small
        \frac{\partial L_{PSD}}{\partial \boldsymbol{v}_i} 
        =\sum_{j}^{\left | \mathcal{B} \right |} \left ( \boldsymbol{z}_j -\left ( \boldsymbol{z}_i \cdot \boldsymbol{z}_j \right ) \boldsymbol{z}_i\right ) \left (   P_{ij} - S_{ij}\right )
    \label{appeq:final_grad_real}
\end{equation}
which is equivalent to Eq \textcolor{red}{8}.

\noindent \textbf{Proposition 2.} Eq \textcolor{red}{5} is equivalent to the optimization problem Eq \textcolor{red}{11}.

\noindent \textit{Proof.} We have Eq \textcolor{red}{4}:
\begin{equation}
\small
        A = (1-\omega)(I-\omega S)^{-1}D
    \label{appeq:diffusion}
\end{equation}
Both sides of the Eq~\ref{appeq:diffusion} multiplied by $(I-\omega S)$, then move the right side to the left side, we get:
\begin{equation}
\small
        (I-\omega S)A-(1-\omega)D = 0
    \label{appeq:diffusion_prop1}
\end{equation}
Eq~\ref{appeq:diffusion_prop1} can also be transformed as:
\begin{equation}
\small
        (I-S)A+\frac{1-\omega}{\omega}(A-D) = 0
    \label{appeq:diffusion_prop2}
\end{equation}
Define $\tilde{A}$ as the vectorization of matrix $A$. The vectorization of a matrix is a linear transformation that vectorizes an input matrix by stacking its columns one after the next. Vectorize the both sides of Eq~\ref{appeq:diffusion_prop2}, we get:
\begin{equation}
\small
        (I\otimes(I-S))\tilde{A}+\frac{1-\omega}{\omega}(\tilde{A}-\tilde{D}) = 0
    \label{appeq:diffusion_prop3}
\end{equation}
where $\otimes$ indicates the Kronecker Product. Since $\tilde{A}$ and $\tilde{A}$ are vectors and $(I\otimes(I-S))$ is a symmetric matrix, we can see Eq~\ref{appeq:diffusion_prop3} as the partial derivative of an objective function, the objective function can be described as:
\begin{equation}
\small
        \min _{\tilde{A}} \tilde{A}^{\mathrm{T}}(I\otimes(I-S))\tilde{A}+\frac{1-\omega}{\omega}\|\tilde{A}-\tilde{D}\|_2^2
    \label{appeq:diffusion_prop4}
\end{equation}
Define $Y \equiv |\mathcal{B}|(j-1)+m$ and $Z \equiv |\mathcal{B}|(k-1)+n$, the first term can be described as:
\begin{equation}
\small
    \begin{aligned}
        &\tilde{A}^{\mathrm{T}}(I\otimes(I-S))\tilde{A}  \\
        =& \tilde{A}^{\mathrm{T}}(I\otimes(I-V^{-1/2}WV^{-1/2}))\tilde{A} \\
        =& \sum_{Y=1}^{|\mathcal{B}|^2} \tilde{A}_Y^2 - \tilde{A}^{\mathrm{T}}(I\otimes(V^{-1/2}WV^{-1/2}))\tilde{A} \\
        =& \sum_{Y,Z=1}^{|\mathcal{B}|^2} (I\otimes W)_{Y,Z}\frac{\tilde{A}_Y^2}{(I\otimes V)_{Y,Y}} \\
        &- \sum_{Y,Z=1}^{|\mathcal{B}|^2} \tilde{A}^{\mathrm{T}}\frac{(I\otimes W)_{Y,Z}}{\sqrt{(I\otimes V)_{Y,Y}(I\otimes V)_{Z,Z}}}\tilde{A} \\
        =& \frac{1}{2}\sum_{Y,Z=1}^{|\mathcal{B}|^2}(I\otimes W)_{Y,Z}(\frac{\tilde{A}_{Y}}{\sqrt{(I\otimes V)_{Y,Y}}}-\frac{\tilde{A}_{Z}}{\sqrt{(I\otimes V)_{Z,Z}}})^2 \\
        =& \frac{1}{2} \!\sum_{i, j, k=1}^{|\mathcal{B}|} \!W_{j,k}\!\left(\frac{A_{i,j}}{\sqrt{V_{i,i} V_{j,j}}}-\frac{A_{i,k}}{\sqrt{V_{i,i} V_{k,k}}}\right)^2
    \end{aligned}
\end{equation}
And the second term can be described as:
\begin{equation}
\small
    \frac{1-\omega}{\omega} \sum_{i,j=1}^{|\mathcal{B}|}\left(A_{i,j}-D_{i,j}\right)^2
    = \frac{1-\omega}{\omega}\|\tilde{A}-\tilde{D}\|_2^2
\end{equation}
So Eq~\ref{appeq:diffusion_prop4} can be transformed to:

\begin{equation}
\small
\begin{split}
    &\min _A \frac{1}{2} \!\sum_{i, j, k=1}^{|\mathcal{B}|} \!W_{j,k}\!\left(\frac{A_{i,j}}{\sqrt{V_{i,i} V_{j,j}}}-\frac{A_{i,k}}{\sqrt{V_{i,i} V_{k,k}}}\right)^2\\
    &+\frac{1-\omega}{\omega} \sum_{i,j=1}^{|\mathcal{B}|}\left(A_{i,j}-D_{i,j}\right)^2
\end{split}
\end{equation}
which is equivalent to Eq \textcolor{red}{11}.

\section{Ablation Studies on Hyper-parameters}
In this section, we investigate the effect of the distillation weight $\lambda$ in Eq \textcolor{red}{2} and transition probability of random walk $\omega$ in Eq \textcolor{red}{5}. To investigate their effect, we provide the validation performances in terms of Recall Top-1 (R@1) on CUB200-2011, CARS196, and Stanford Online Products datasets with Multisimilarity loss and R50/128, and follow the setting described in Section \textcolor{red}{5.1} of the main paper. 

\begin{figure}[ht]
  \centering
    \includegraphics[width=0.9\linewidth]{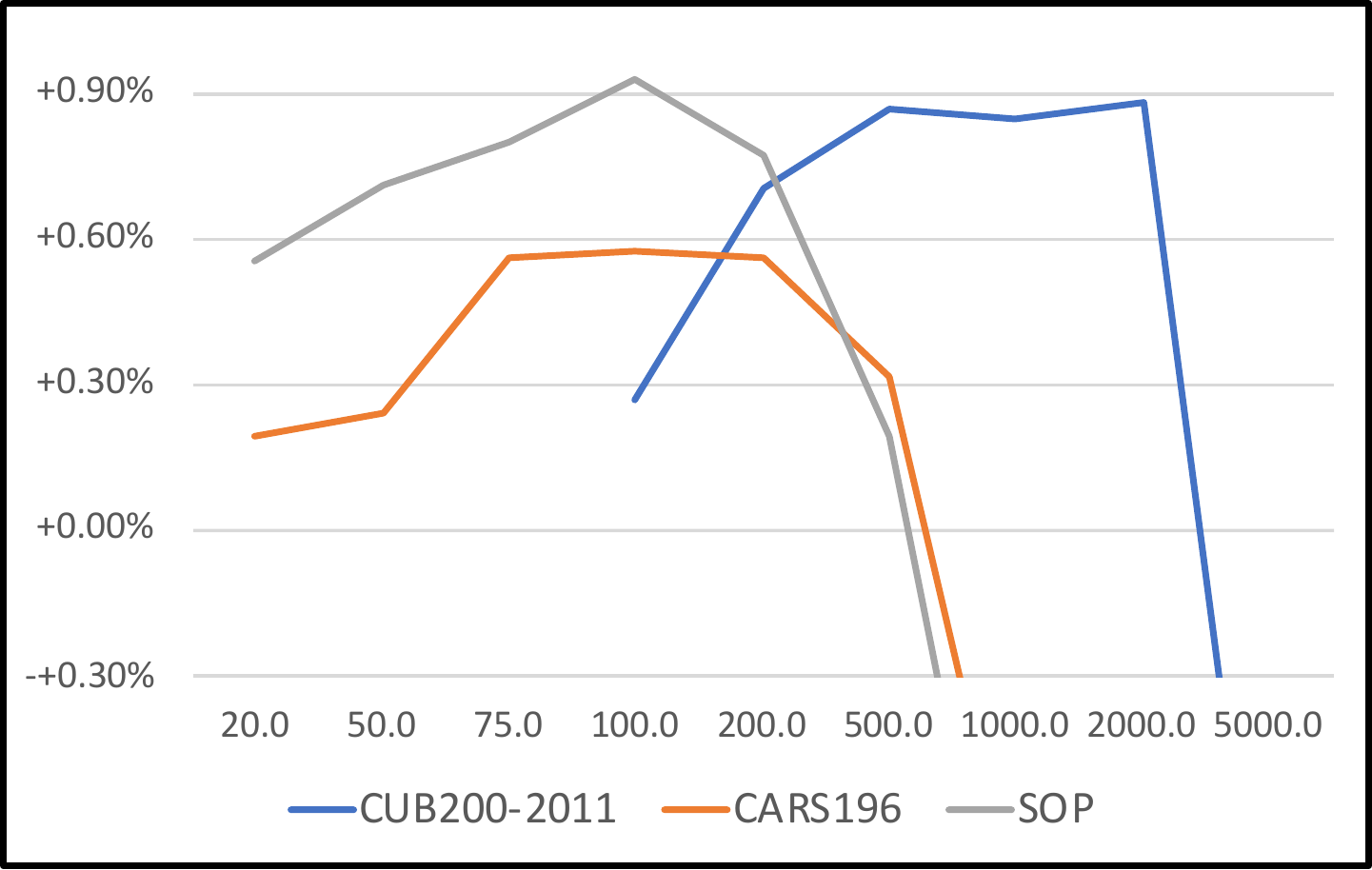}
    \caption{The effect of the distillation weight $\lambda$.}
  \label{fig:aplha}
\end{figure}

\noindent \textbf{The effect of distillation weight $\lambda$.} 
Figure~\ref{fig:aplha} shows the results. The Y-axis describes the performance improvement relative to the baseline and X-axis indicates the value of $\lambda$. For CUB200-2011, the performance is very similar when $\alpha \in [500,2000]$, so we set $\alpha=1000$. For the same reason, we set $\alpha=75$ for CARS196. For SOP, we set $\alpha=100$ since the improvement achieves the peak around $\alpha=100$. 

\noindent \textbf{The effect of transition probability of random walk $\omega$.} As demonstrated in Table~\ref{tab:omega} the performance is consistent when changing $\omega$ from 0.1 to 0.99, with peaks around $\omega=0.3$, $\omega=0.99$ for CUB200-2011 and CARS196. For Stanford Online Products, since the performance is very similar when $\omega \in [0.5,0.99]$, we set $\omega=0.5$.

\begin{table}[ht]
  \caption{The effect of transition probability of random walk $\omega$..}
  \tabcolsep=10pt
  \centering
  \resizebox{1.0\linewidth}{!}{
  \begin{tabular}{lccc}
    \toprule
    Datasets$\rightarrow$ & CUB200-2011 & CARS19& SOP \\
    \midrule
       &\textbf{R@1}    &\textbf{R@1}    &\textbf{R@1} \\
    \midrule
    PSD w/o OBDP &63.46  &82.34   &78.53   \\
    \midrule
    $\omega=0.1$ &65.19  &82.83   &78.66   \\
    $\omega=0.3$ &\textbf{67.76}  &82.82   &78.77   \\
    $\omega=0.5$ &67.56  &82.88   &\textbf{78.83}   \\
    $\omega=0.7$ &67.19  &83.32   &\textbf{78.85}   \\
    $\omega=0.9$ &66.90  &\textbf{83.66}   &\textbf{78.84}   \\
    $\omega=0.99$ &66.94  &84.10   &\textbf{78.84}   \\
    \bottomrule
  \end{tabular}
  }
  \label{tab:omega}
\end{table}

\section{Visualization}
To better observe and understand the effectiveness of OBD-SD, we show some visualization results in this section. 

\subsection{Visualization of feature distribution}
\begin{figure*}[t]
\centering
    \begin{tabular}{cc}
        \includegraphics[width=0.5\linewidth]{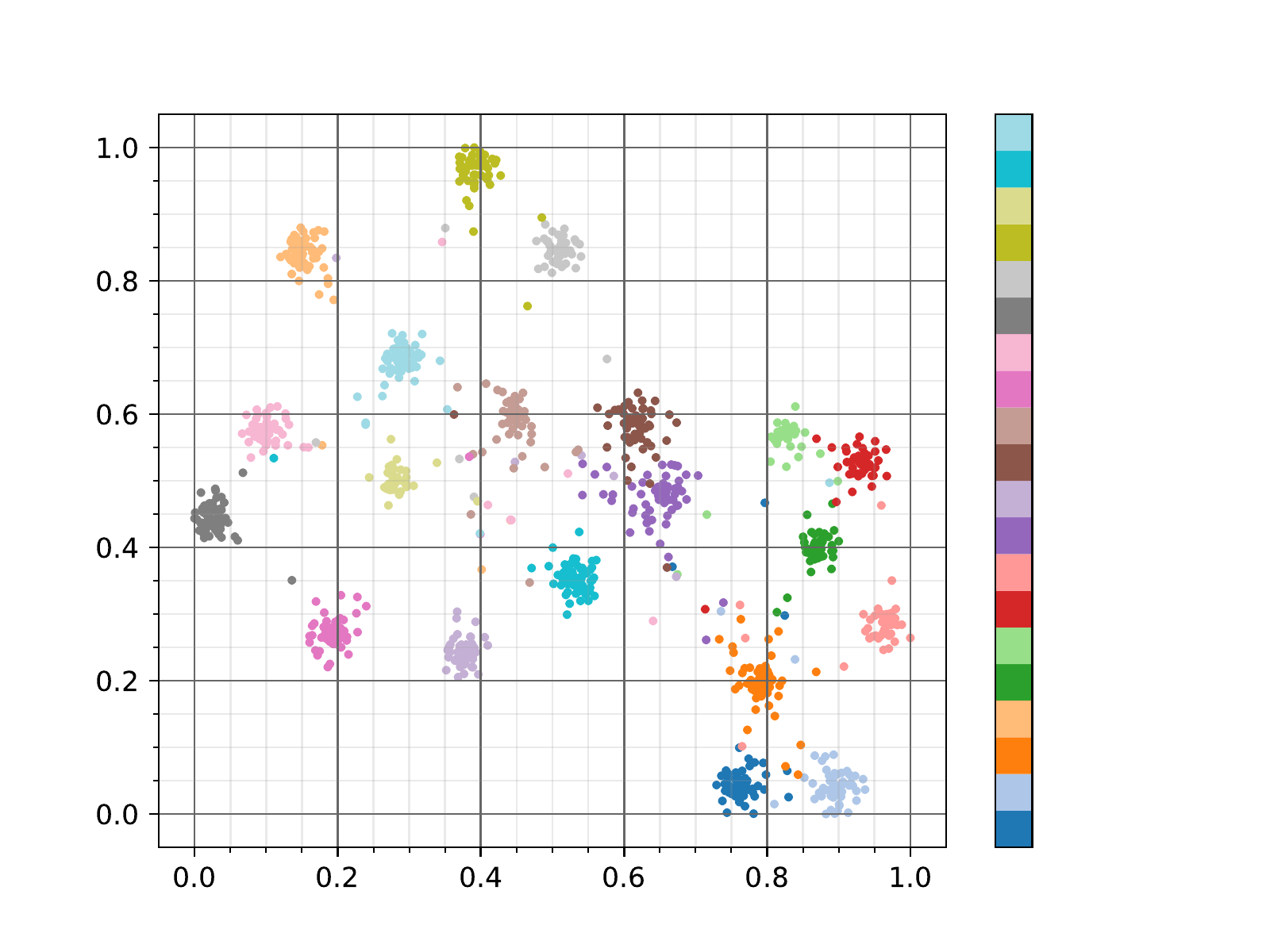} &
        \includegraphics[width=0.5\linewidth]{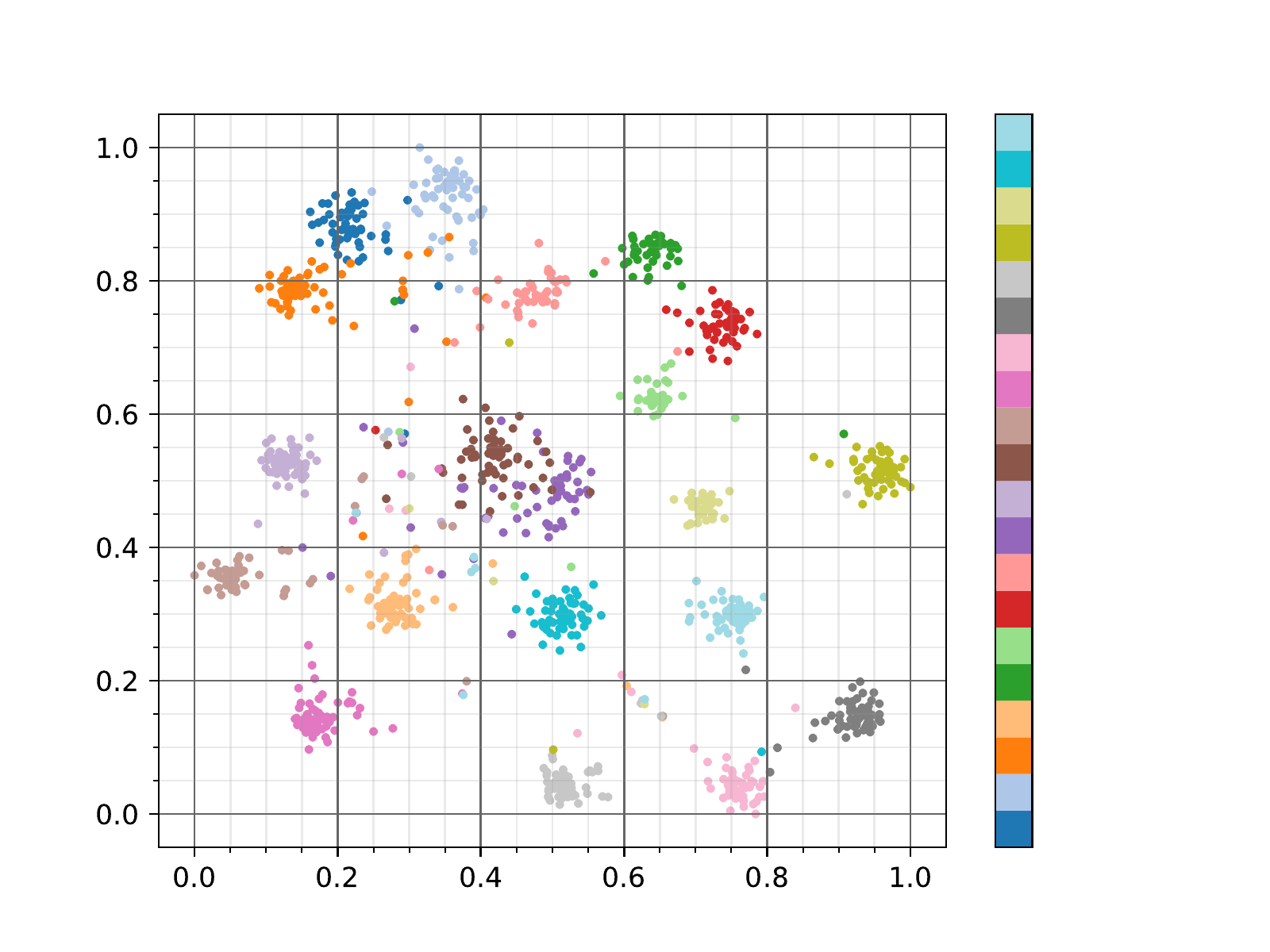} \\
        \small{(a) MS w/o OBD-SD on CUB200-2011}    &
        \small{(b) MS with OBD-SD on CUB200-2011}    \\
        \includegraphics[width=0.5\linewidth]{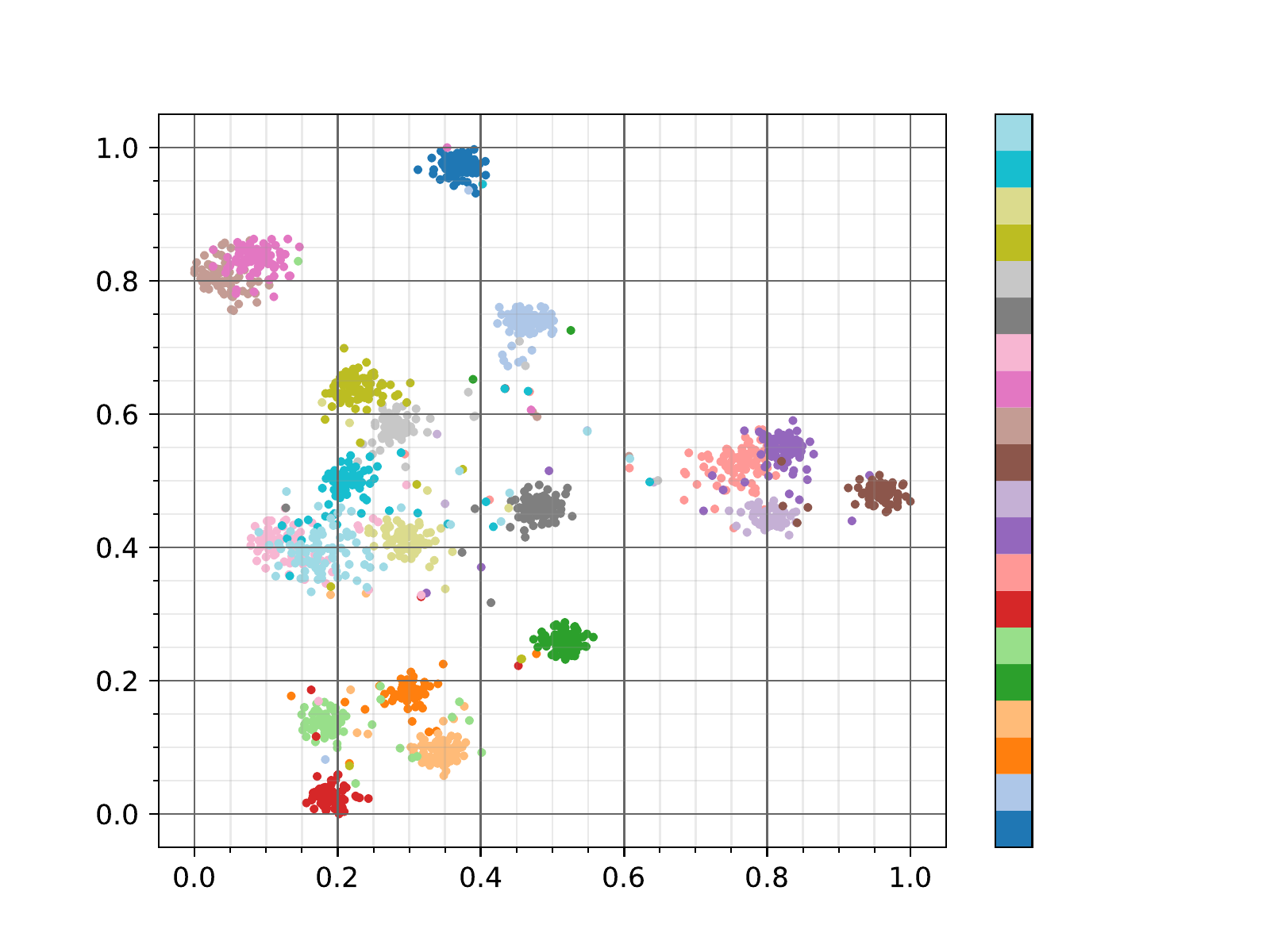} &
        \includegraphics[width=0.5\linewidth]{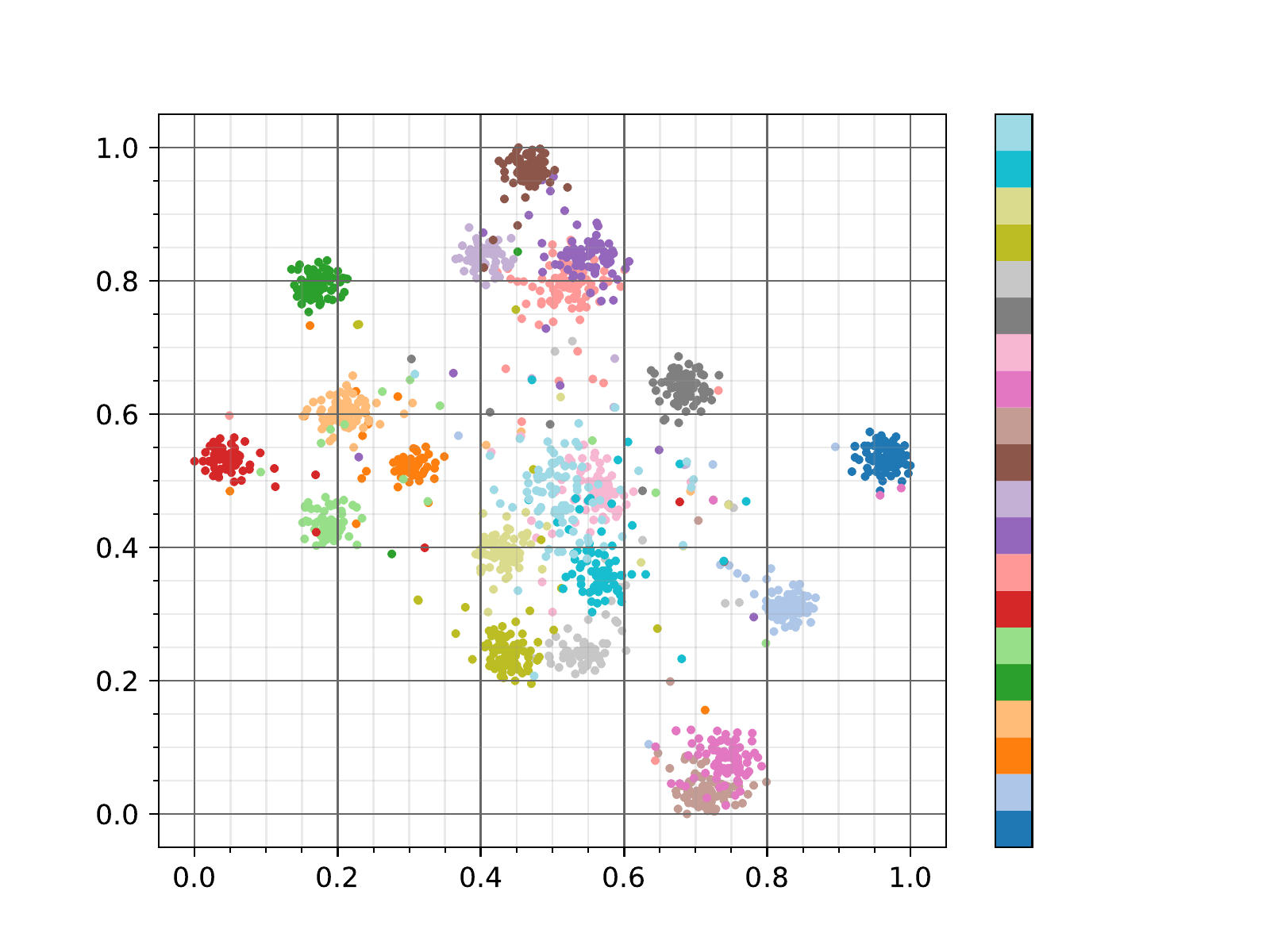} \\ 
        \small{(c) MS w/o OBD-SD on CARS196}    &
        \small{(d) MS with OBD-SD on CARS196}    \\
   \end{tabular}
       \caption{Visualization of feature distributions. We use R50/128 as backbone and MS loss to train all models. (a) and (b) show the feature distributions of models trained with and without OBD-SD on the CUB200-2011 dataset, respectively. (c) and (d) show the feature distributions of models trained with and without OBD-SD on the CARS196 dataset, respectively. The colorbar shows 20 different colors, each referring to a different class.}
\label{fig:visualization_space}
\end{figure*}

To observe the effect of OBD-SD on the embedding space more visually, we visualized the feature distributions of the first 20 classes of samples from the training subset of CUB200-2011 and CARS196 datasets using t-SNE. Fig~\ref{fig:visualization_space} shows the results. We see that OBD-SD makes the intra-class distribution more decentralized while maintaining a clear inter-class boundary. In other words, OBD-SD can reduce over-clustering and encourage a smoother spatial distribution, which is better for the generalization of the embedding space. This conclusion is consistent with our analysis in Section \textcolor{red}{5.6} of the main paper and also with our motivation.

\subsection{Visualization of image retrieval results}
\begin{figure*}[ht]
\centering
\includegraphics[width=1.0\linewidth]{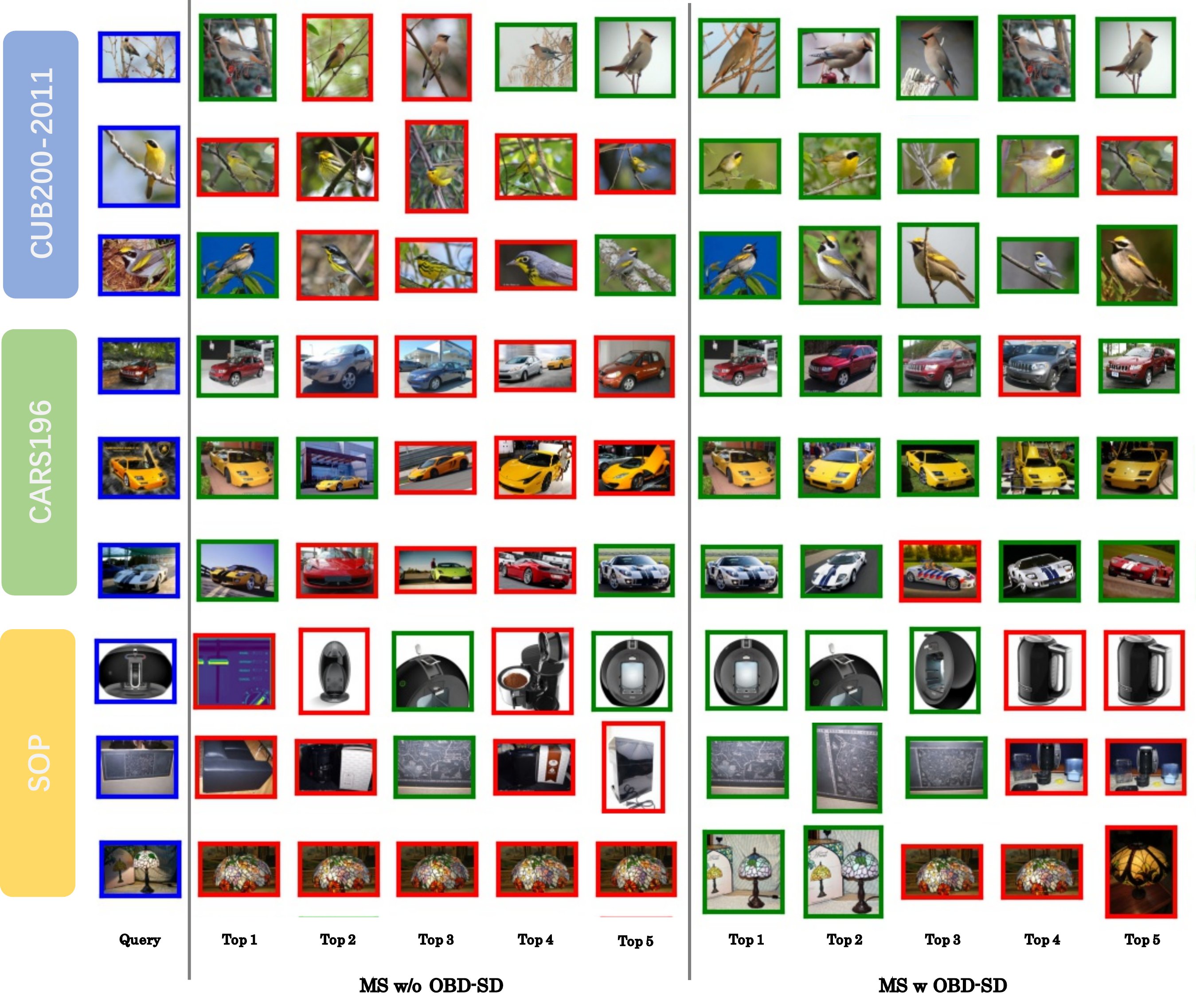}
\caption{The visualization of top 5 retrieved images w/o and w/ OBD-SD on three datasets. Green and red borders indicate correct and incorrect retrieved results, respectively.}
\label{fig:ranking_visualization}
\end{figure*}

we visualize the image retrieval results on the test set of the three datasets to investigate the capability of OBD-SD. Specifically, given a query image, we display its top 5 retrieved gallery images. The visualization results are shown in Fig.~\ref{fig:ranking_visualization}. We see that OBD-SD can improve the ability to represent fine-grained information in the embedding space.
For example, in the third row of the CUB200-2011 part, the model trained with OBD-SD noticed subtle differences in the pattern of the species from other similar birds. In the second row of the CARS196 part, OBD-SD helps the model notice the difference in the headlights. 
These are attributed to the fact that OBD-SD provides richer and more accurate information about the intrinsic relationships of the samples during the training.

\begin{figure*}[ht]
  \centering
  \begin{subfigure}{1.0\linewidth}
    \includegraphics[width=1.0\linewidth]{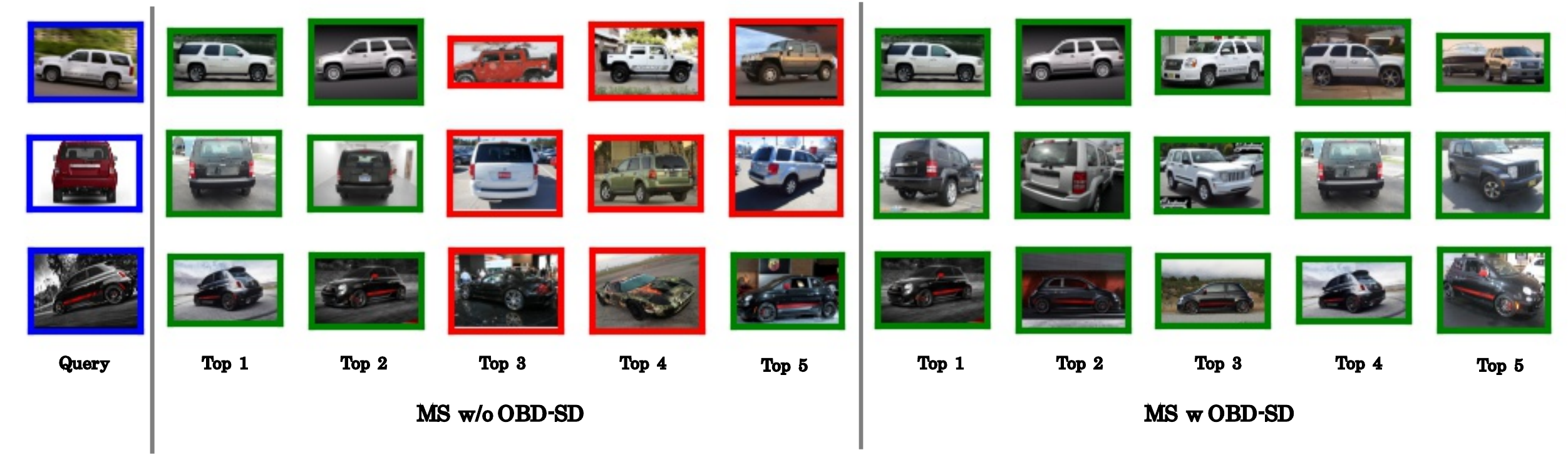}
    \caption{Cross-view image retrieval.}
    \label{fig:short-a}
  \end{subfigure}
  \begin{subfigure}{1.0\linewidth}
    \includegraphics[width=1.0\linewidth]{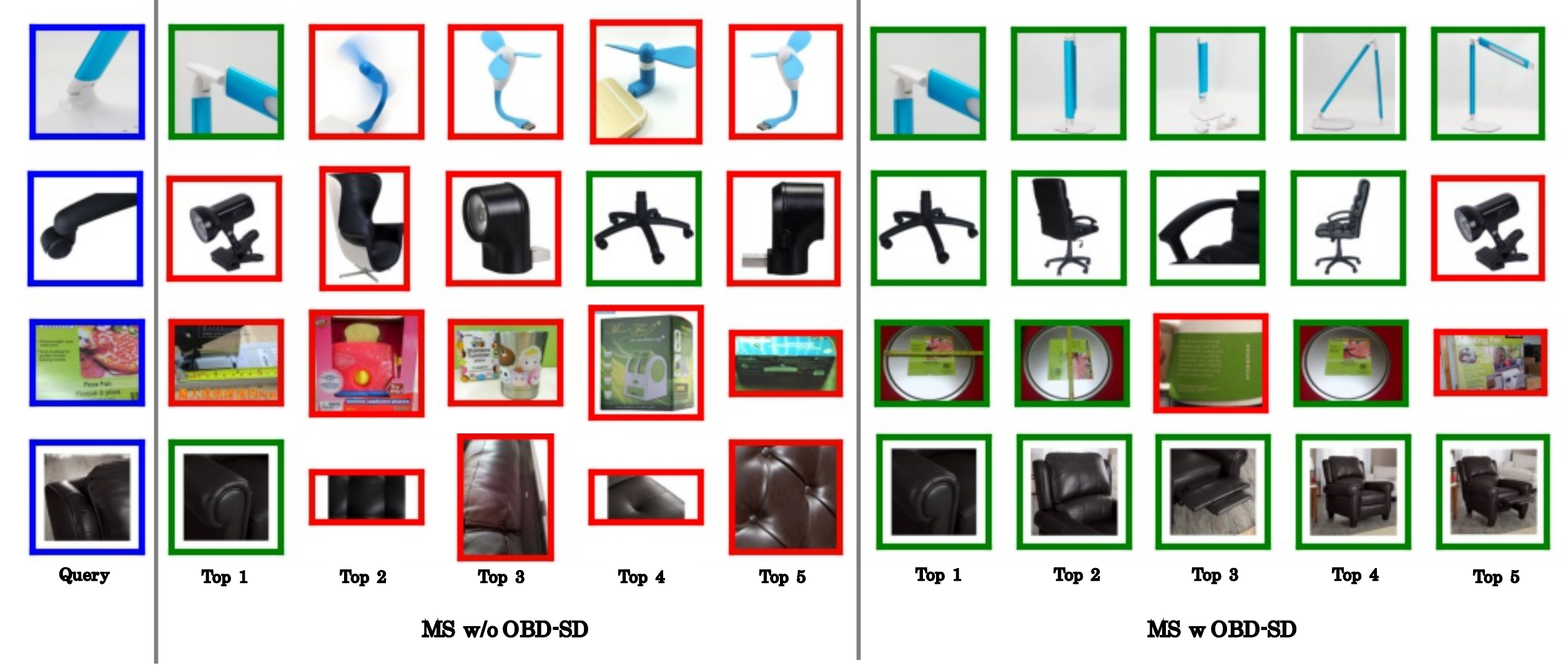}
    \caption{Local-to-global image retrieval..}
    \label{fig:short-b}
  \end{subfigure}
  \caption{The visualization of top 5 retrieved images w/o and w/ OBD-SD on three datasets for (a) cross-view image retrieval and (b) local-to-global image retrieval. Green and red borders indicate correct and incorrect retrieved results, respectively. }
  \label{fig:ranking_visualization_more}
\end{figure*}

Moreover, to validate the ability of OBD-SD in two aspects: corss-view and local-to-global, we show more image retrieval results:
\begin{itemize}
    \item \textbf{Cross-view}: The positive samples in the gallery have different views from the query image. Figure~\ref{fig:short-a} shows that the model trained with OBD-SD can work well on the cross-view task. For example, in the second row, the model can retrieve the corresponding car's gallery images, even including the front-view images, based on the rear-view query.
    \item \textbf{Local-to-global}: The query image contains only the local content of the target. Figure~\ref{fig:short-b} shows that OBD-SD improves the model's capability to handle ``local-to-global". For example, in the second row, the model trained with OBD-SD is aware of the query image is the wheels of a swivel chair. 
\end{itemize}

\begin{figure*}[ht]
\centering
\includegraphics[width=1.0\linewidth]{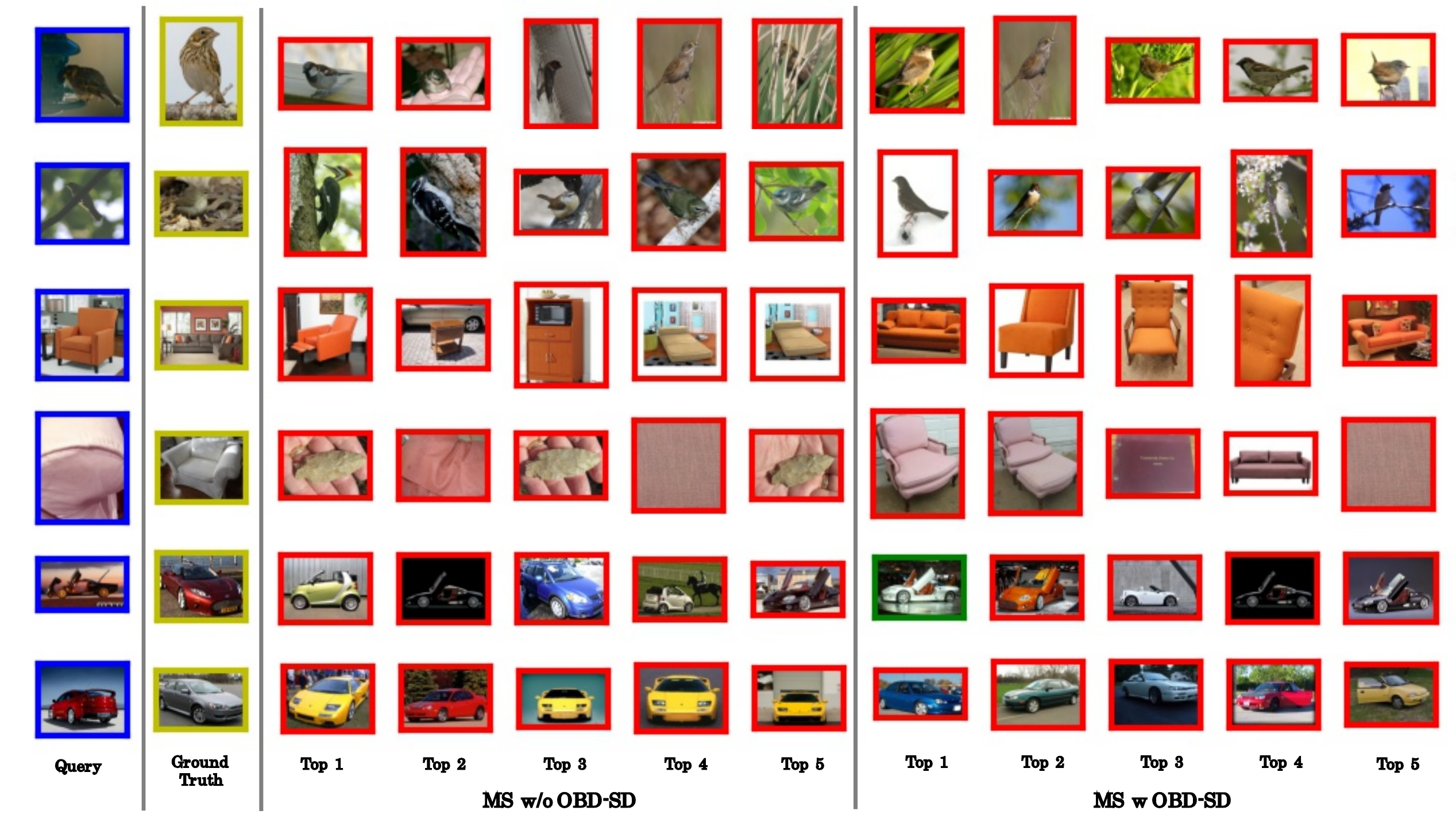}
\caption{Visualization failure cases. Given a query image (blue), we present a ground truth sample (yellow) and the visualization of the top 5 retrieved images w/o and w/ OBD-SD. Green and red borders indicate correct and incorrect retrieved results, respectively.}
\label{fig:bad}
\end{figure*}

Finally, we also present some bad cases for those very difficult samples. The results are shown in Figure~\ref{fig:bad}, where retrieved results of OBD-SD are visually closer to the ground truth and query than that of the baseline method.

\section{Code}
We also share the code and some pre-trained models to validate our paper’s results are reproducible and trustworthy. The project of the code in the folder namely ``OBD-SD\_Pytorch-main". Please see the README.md in the project for more information.

{\small
\bibliographystyle{ieee_fullname}
\bibliography{Supp}
}